# Simultaneous Localization and Mapping Related Datasets: A Comprehensive Survey

Yuanzhi Liu, Yujia Fu, Fendong Chen, Bart Goossens, Wei Tao, and Hui Zhao

*Abstract*—**Due to the complicated procedure and costly hardware, Simultaneous Localization and Mapping (SLAM) has been heavily dependent on public datasets for drill and evaluation, leading to many impressive demos and good benchmark scores. However, with a huge contrast, SLAM is still struggling on the way towards mature deployment, which sounds a warning: some of the datasets are overexposed, causing biased usage and evaluation. This raises the problem on how to comprehensively access the existing datasets and correctly select them. Moreover, limitations do exist in current datasets, then how to build new ones and which directions to go? Nevertheless, a comprehensive survey which can tackle the above issues does not exist yet, while urgently demanded by the community. To fill the gap, this paper strives to cover a range of cohesive topics about SLAM related datasets, including general collection methodology and fundamental characteristic dimensions, SLAM related tasks taxonomy and datasets categorization, introduction of state-of-the-arts, overview and comparison of existing datasets, review of evaluation criteria, and analyses and discussions about current limitations and future directions, looking forward to not only guiding the dataset selection, but also promoting the dataset research.**

## I. Introduction

Simultaneous Localization and Mapping (SLAM) is a subject with respect to achieving both "internal and external cultivation" in a mobile robot – externally knowing the world by constructing a map and internally knowing itself by tracking the location [1], [2], [3]. Serving as the decisive technology of mobile robotic community, SLAM has attracted significant attention and aroused many productive applications. In self-driving cars [4], navigating robots [5], and unmanned aerial vehicles (UAVs) [6], [7], SLAM maps the surrounding environment and figures out the ego-motion inside the map. In virtual reality/augmented reality (VR/AR) systems [8], SLAM gives accurate poses of the agent, thus helps to intuitively manipulate virtual objects. In mobile mapping [9], SLAM estimates transformations across adjacent frames and merges together the 3D points into a consistent model. However, from the real-world deployment point of view, such technologies are far from mature, indicating a huge gap between benchmark scores and mature productivity [10], [11].

The research of SLAM related problems has been heavily relying on publicly available datasets. On the one hand, the concrete experiment of SLAM related problems involves quite costly hardware and complicated procedure. One has to employ mobile platforms (such as a wheeled robot), sensors (such as LiDARs and cameras), and ground truth instruments (such as a real-time kinematic (RTK) corrected global navigation satellite system (GNSS) and a set of motion capture system), which are huge financial burdens for individual researchers. And if multi-sensor data are required (in most cases it is), the temporal and spatial calibration is obligatory to be performed before data collection. Overall, it is almost impossible to test every draft in the manner of field experiment. On the other hand, datasets can provide equitable benchmarks for evaluation and comparison. To fairly quantify the algorithm performance, the dataset providers have proposed several consensual evaluation criteria and indicators [12], [13]. By evaluating on the same datasets, researchers worldwide are able to compare their algorithms horizontally. Gradually, public benchmark score has become a fundamental basis of today's paper publication.

In this context, four major problems arose. Firstly, the usages of existing datasets are quite biased. Actually, over the past decade, a quantity of datasets have been released, whereas the researchers lack a convenient way to know them, thus only test on several best-known ones, such as KITTI [14], TUM RGB-D [12], and EuRoC [15]. This may well result in over-fitting, thus good scores on a certain benchmark cannot ensure equally good results on the others. Secondly, the selection of datasets lacks a clue and is sometimes incorrect. SLAM is an emerging but complicated subject, while still there is no clear taxonomy on related tasks. This causes a two-way selection difficulty: the algorithm researchers have no clue to choose suitable datasets, and the dataset providers have no basis to specify the evaluable tasks of them. Thirdly, the existing datasets are still limited in some dimensions. For example, lack of scene types, data size, and sensor number and modalities, limitation of ground truth, etc. Although different datasets are complementary, still the datasets in their entirety do not reflect all-roundly the real world, thus could easily hide the defects of algorithms. Nevertheless, such limitations are rarely analyzed to recommend future direc-

This work is supported by the National Key R&D Program of China under Grant 2018YFB1305005.

Yuanzhi Liu, Yujia Fu, Wei Tao, and Hui Zhao are with the Department of Instrument Science and Engineering, School of Electronic Information and Electrical Engineering, Shanghai Jiao Tong University, Shanghai 200240, China (e-mail: lyzrose@sjtu.edu.cn, yujiafu@sjtu.edu.cn, taowei@sjtu.edu.cn, huizhao@sjtu.edu.cn).

Fengdong Chen is with the School of Instrumentation Science and Engineering, Harbin Institute of Technology, Harbin 150001, China (e-mail: chenfd@hit.edu.cn).

Bart Goossens is with the Department of Telecommunications and Information Processing (TELIN), Ghent University, 9000 Ghent, Belgium, and also with the Interuniversity Microelectronics Center (imec), 3001 Leuven, Belgium (e-mail: bart.goossens@ugent.be).

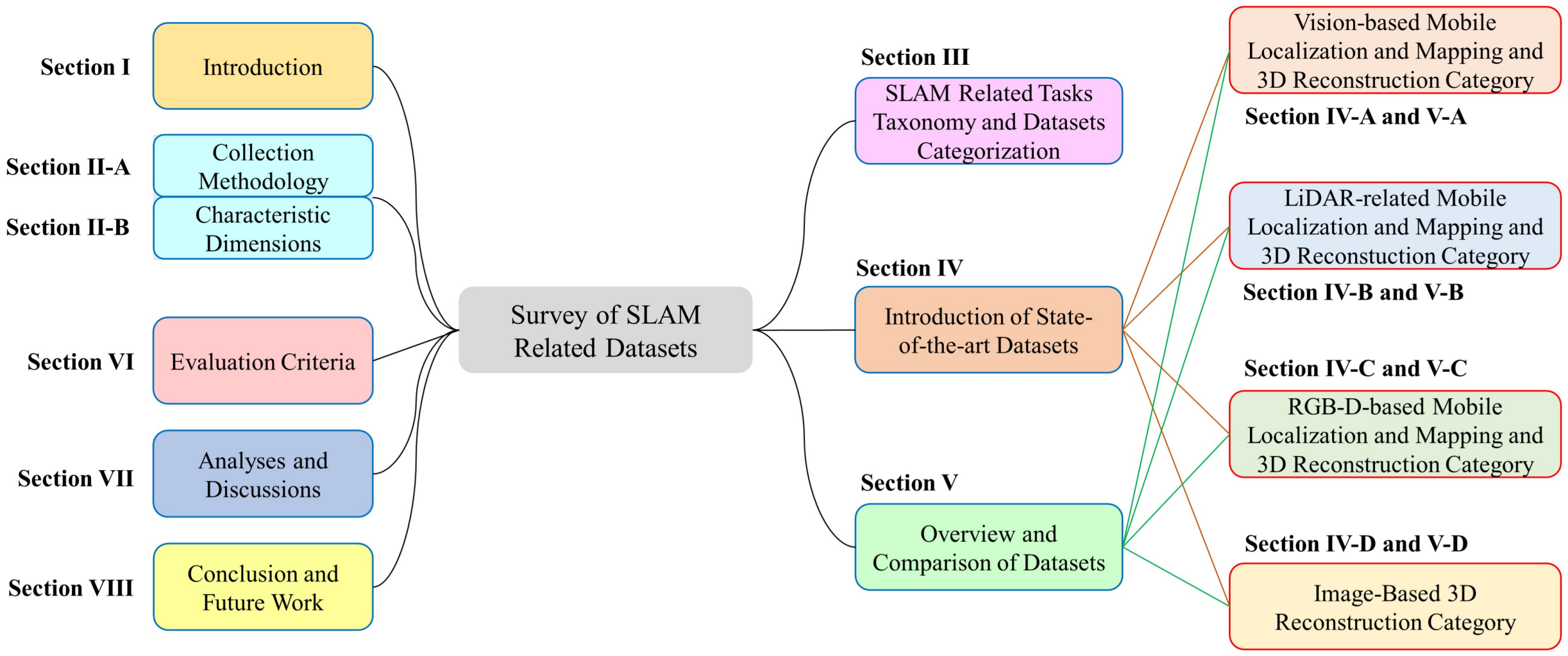

Fig. 1. The schematic diagram of the overall structure of the paper.

tions. Fourthly, the threshold in creating datasets is relatively high. The research of SLAM related datasets is a self-contained subject which can push the limits of algorithms at another level, while still the community cannot access a systematical knowledge about collection methodology and characteristics, which may well affect the widespread releases of new datasets.

To the best of our knowledge, no open literature has addressed the above issues. Even, there is not any survey specializing in SLAM datasets yet. Firman [16] surveyed a total of 102 RGB-D datasets, while out of them, only 9 can be used for SLAM related problems. Cai et al. [17] also surveyed the RGB-D-based datasets, and 46 datasets were categorized into 5 categories, whereas covered only 2 for SLAM usages. Bresson et al. [18] presented a survey about the trends of SLAM in autonomous driving with 10 datasets summarized, nevertheless, they were quite old and mainly collected for LiDAR-based methods, which were far from comprehensive. Yurtsever et al. [19] also presented a survey about autonomous driving, with an overview of 18 datasets. However, the applicable tasks were not specified: many of them cannot be used for SLAM purpose.

To fill the gap, this paper strives to present a structured and comprehensive survey about SLAM related datasets. On the one hand, as there is not any perfect or dominant dataset in SLAM domain over this period (as ImageNet [20] in computer vision field), we believe SLAM related datasets should be used in a manner of complementation. To this end, the primary goal of this paper is to serve as a dictionary to ensure comprehensive, correct, and efficient dataset selections. Based on the proposed SLAM related tasks taxonomy and datasets categorization, we survey a total of 97 datasets with every dataset specified with the applicable tasks, and meanwhile, many dimensions of characteristics are summarized and compared. On the other hand, whether from the perspective of drilling or evaluation, we believe the datasets can push the limits of SLAM related algorithms in another level. Thus, another goal of this paper is to lower the threshold and guide the future directions of dataset research. Therefore, the general collection methodology and fundamental characteristic dimensions of SLAM related datasets are described, and the current limitations and future directions are analyzed and discussed. Moreover, some state-of-the-art datasets and consensual evaluation criteria are introduced in detail, which are instructive to future releases.

The remainder of this paper is organized into 5 sections (the overall structure is shown in Fig. 1). Section II introduces the general collection methodology and fundamental characteristic dimensions of SLAM related datasets, in order to pave the way for the following introductions. Section III clears the taxonomy of SLAM related tasks, resulting in four broad categories for datasets classification. Section IV introduces 16 state-of-the-art datasets in detail. Section V surveys by category as many as possible existing SLAM related datasets (97 in total), forming into 2 overview and comparison tables for each category, with guidelines on datasets selection. Section VI reviews the key evaluation criteria. Section VII gives analyses and discussions on the current weaknesses and future directions of datasets and evaluation. Finally, Section VIII draws a brief conclusion and arranges the future work.

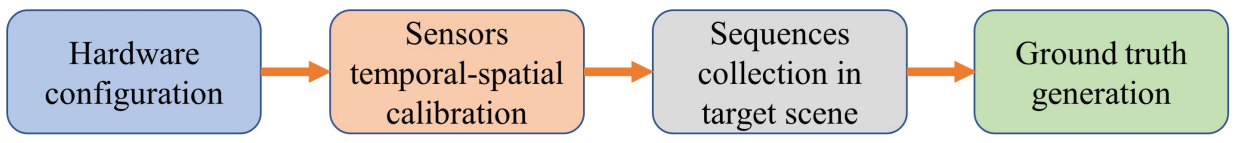

Fig. 2. The general collection methodology of dataset collection.

## II. SLAM Related Datasets: General Collection Methodology and Fundamental Dimensions

### A. *General Collection Methodology*

The general methodology of dataset collection is shown in Fig. 2: configure the hardware first, then perform temporal-spatial sensor calibration, then collect data sequences in a target testing scene, finally generate accurate ground truth. As can be seen, the procedure is more complicated than the other kinds of dataset, such as object detection, which does not require strictly sensor calibration or high-end ground truth generation techniques. To lower the threshold of the research, and pave the way for the following introductions, this part will give a concrete description of the general methodology. It should be noted that, in addition to the real-world data collection, there is also a virtual and synthetic manner, just like the rendering of some realistic 3D computer games. However, the methodology of such manner is mainly a software issue (such as using AirSim[1] and POVRAY[2]), hence we will not introduce it in detail.

*1) Hardware Configuration*

Hardware configuration involves the selection of mobile platforms and perception sensors. For the mobile platforms, the selection mainly depends on the target applications, for example: using a handheld carrier for VR/AR and smartphone applications, a backpack for mobile mapping, an automobile for autonomous driving, a wheeled robot for service and navigation robotics, and an UAV for aerial applications.

For the perception sensors, the selection is required by the sensor type that the algorithm is based on. However, one is quite possible to try different modalities and fusions of sensors to fill a certain gap. Therefore, to support such demands, the collection systems are typically equipped with non-single sensors and multiple modalities, such as cameras (mono, stereo, grayscale, color, and event-based [21]), LiDARs (single- and multi-beam), RGB-D sensor [22], and Inertial Measurement Unit (IMU). Here we mainly refer to commercial-level IMU since it is cheap and ubiquitous, while some high-end IMUs (such as a fiber-optic gyroscope (FOG)) are too expensive and are mainly used as ground truth instruments rather than perception sensors.

*2) Sensors Calibration*

Under the context of multiple sensors configuration (in most cases it is), spatial-temporal calibration among different sensors is obligatory to be performed.

The spatial calibration aims to find a rigid transformation (also called extrinsic parameter) between different sensor frames, enabling to unify all the data into a single coordinate system. Usually, extrinsic parameter calibration may appear in pairs of camera-to-camera, camera-to-IMU, camera-to-LiDAR, and can be extended group-by-group in multiple sensors situation. For stereo camera calibration, the most widely used in machine vision field could be Zhang's method [23] proposed in 2000. It is quite flexible as the technique only requires to freely take images at several different orientations of a printed planar checkerboard. The algorithm will detect corner points on the board among each shot image, serving as the constraints of a closed-form solution, and then a set of coarse values will be solved. Finally, the results will be furtherly refined by minimizing the reprojection error. So far, the intrinsic, lens distortion, and extrinsic parameters are calibrated. The algorithm has been implemented in Matlab[3], OpenCV[4], and ROS[5], proved to be highly accurate and reliable, significantly lowering the bound of 3D vision research. For camera-to-IMU calibration, the best-known framework is the Kalibr[6] toolbox [24], [25], [26]. While calibrating, it only requires to wave the visual-inertial sensor suite in front of the calibration pattern. The framework parameterizes both the transformation matrix and time-offset into a unified principled maximum-likelihood estimator, thus by minimizing the total objective errors, the Levenberg-Marquardt (LM) algorithm can estimate all unknown parameters at once. For camera-to-LiDAR calibration, the general methodology is to find the mutual correspondences between 2D images and 3D point clouds, thus compute a rigid transformation between them (an example is shown in Fig. 3). Two commonly used frameworks are the KIT Camera and Range Calibration Toolbox[7] [27] and the Autoware Calibration Toolbox[8] [28], and they respectively present a fully automatic and an interactive implementation.

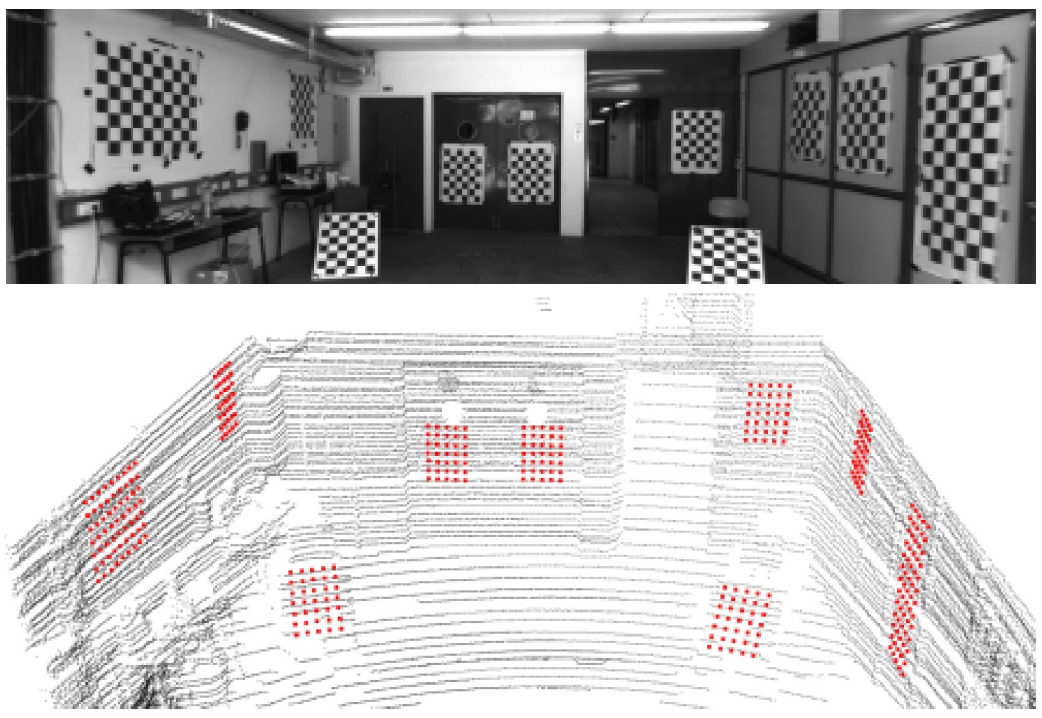

Fig. 3. An example implemented by KIT Range Calibration Toolbox: the top image shows the arrangement of the markers and the bottom shows the result of camera-LiDAR calibration [27].

The temporal calibration aims to synchronize the clocks or data packets of different sensors, or estimate the time-offsets between them. The most desirable solution is to synchronize with the hardware support, such as external trigger (industrial cameras usually have), GNSS timing, and Precision Time Protocol [29] (PTP, also known as IEEE-1588) synchronizing. However, such features can hardly be found on consumer-level products, especially for normal cameras and IMUs. To this end, a feasible approach is to estimate the time-offset caused by unsynchronized clocks or communication delay among different sensors. Solving such problems, the general methodology is to parameterize the time-offset, and furtherly perform optimization by maximizing the cross correlation or minimizing a designed error function between different sensor processes. For example, as mentioned above, the Kalibr toolbox opens an implementation on unified temporal and spatial calibration for camera-to-IMU.

Overall, the procedure and techniques for spatial-temporal alignment among different sensors are quite common, hence we will not detail the calibration process of each dataset in the following sections.

*3) Sequence Collection In Target Scene*

The selection of target scene depends on the expected test scenarios of the algorithms. For example, we can select urban, rural, and highway scenes for autonomous driving test. For data sequence collection, to duplicate the real world better, larger quantity of sequences, various routes, and longer paths are expected. Moreover, if you want to test the possible place recognition and loop closure functions of the algorithm, the route should traverse several times the same location.

[1] https://microsoft.github.io/AirSim/
[2] http://www.povray.org/
[3] https://ww2.mathworks.cn/help/vision/ref/cameracalibrator-app.html
[4] https://docs.opencv.org/master/dc/dbb/tutorial_py_calibration.html
[5] http://wiki.ros.org/camera_calibration
[6] https://github.com/ethz-asl/kalibr
[7] http://www.cvlibs.net/software/calibration/
[8] https://www.autoware.ai/

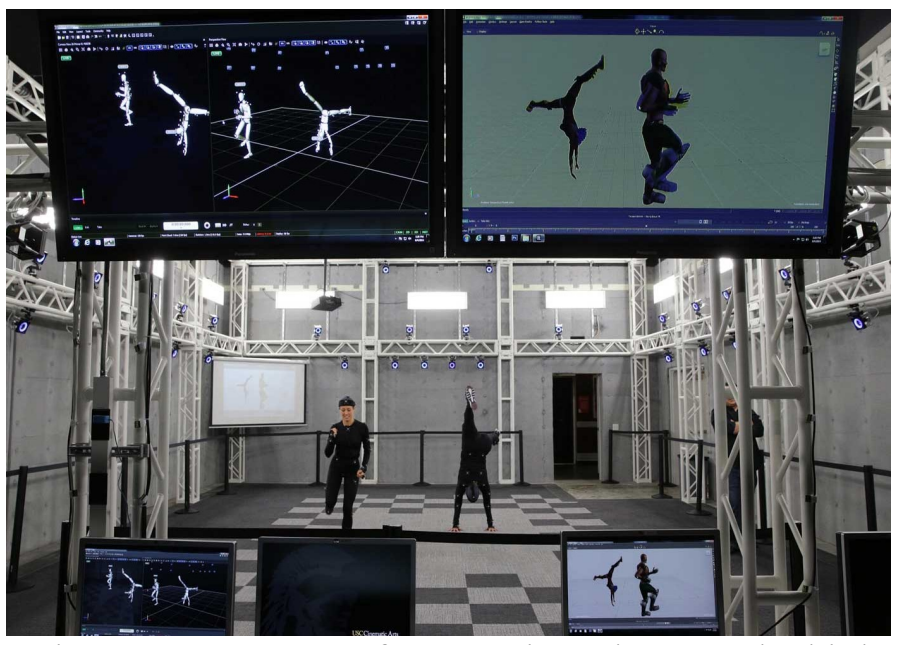

Fig. 4. The motion capture system from OptiTrack [30]. The high-rate cameras mounted around the room are tracking the location and posture of the actors by detecting the reflective markers.

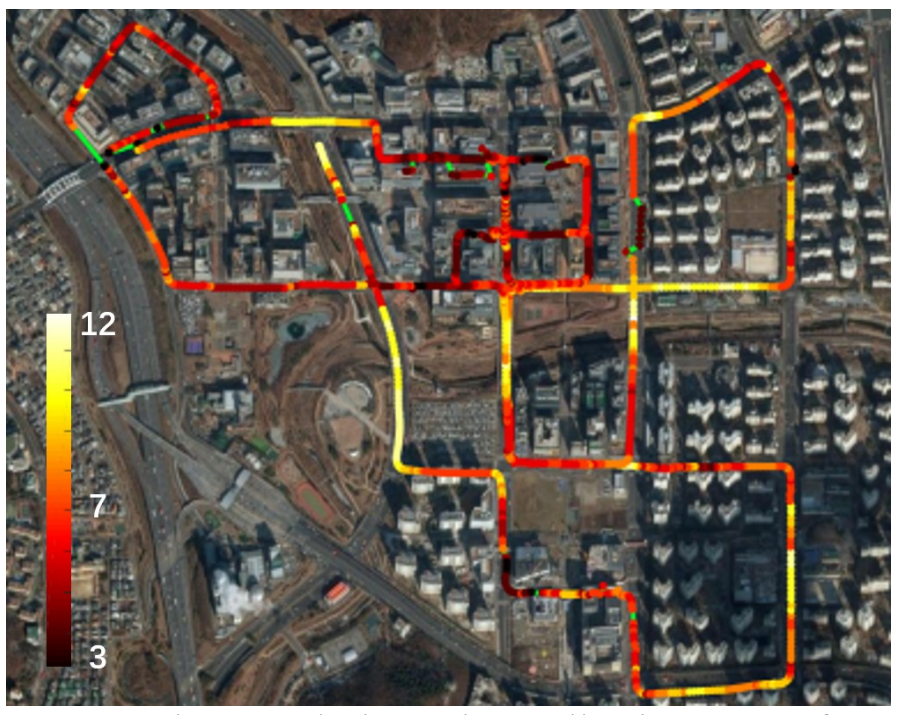

Fig. 5. The GPS reception rate during a data collection route of Complex Urban dataset. Trajectory color represents the number of received satellites: Yellow more, Red fewer, and Green represents invalid. In complex urban area, it is challenging for GNSS to work interrupted by high-rise buildings [39].

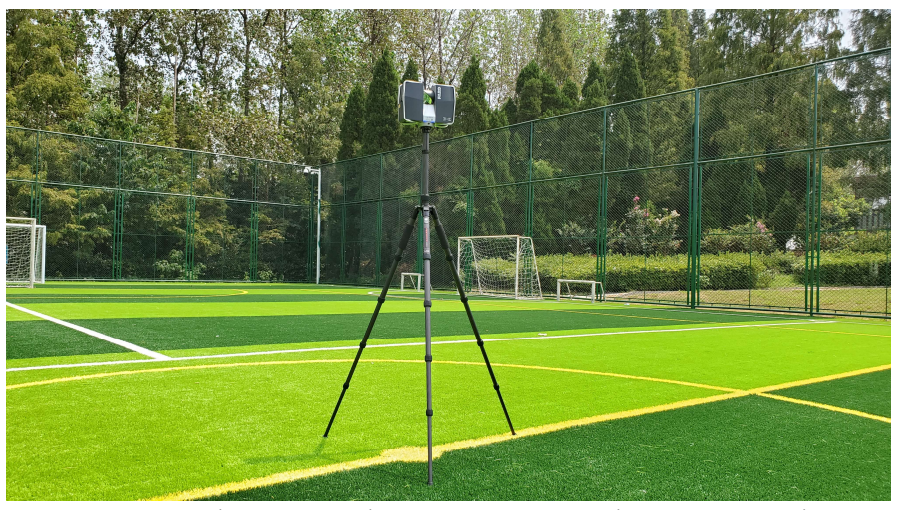

Fig. 6. The FARO X130 large-scale 3D scanner that we used to get the ground truth 3D model of a football court. It scans scene geometry by rotating its laser emitter globally and measuring the return wave to compute each 3D point.

### 4) *Ground Truth Generation*

Ground truth is the accurate reference data to be compared against in evaluation. To make it clear, within the range of SLAM related problems, we only talk about two kinds of ground truth: pose (namely position and orientation) and 3D structure, respectively for localization and mapping evaluation.

- *Position and Orientation*:

The technique employed for pose measurement could differ regarding different scenarios. For small-scale and indoor scenes, the most effective solution is motion capture system (MoCap) (e.g., VICON[9], OptiTrack[10], and MotionAnalysis[11]) (as shown in Fig. 4), which works at a rather high rate, and meantime provides incredible accuracy both on position and orientation. But there are still drawbacks of it: it is quite costly and cannot work well outdoors, as the excessive interference of infrared light will affect the camera tracking. For large-scale outdoor scenarios, GNSS with RTK correction and dual-antenna could be the most mature solution [31]. However, the applicable condition is also strict – the mobile agent should be under an open sky, otherwise the accuracy will drift significantly. So quite commonly, inside a single data sequence, there could be some segments that cannot be trusted, which will dramatically influence the quality of the dataset (as shown in Fig. 5). Another high-precision solution supporting both indoor and outdoor environment is laser tracker (e.g., FARO[12] and Leica[13]). It provides with superior positioning accuracy (up to sub-mm level), while occlusion-free should be ensured to make way for the laser emitted from instrument to the target, which is hard to achieve in many situations. There are also some alternatives for localization, such as Fiducial Marker [32], Global System for Mobile Communications (GSM) [33], ultra-wide band (UWB) [34], Wi-Fi (we refer to the wireless network technology based on the IEEE 802.11 family of standards) [35], and Bluetooth [36], all of which are widely deployed in industry. However, although more ubiquitous, these solutions are either with insufficient accuracy or struggle in certain conditions. Moreover, sometimes the environments could be too complex to measure ground truth (e.g., urban canyon and indoor-outdoor integration scene), then it is a compromise to use the results of multi-sensor fusion algorithms (e.g., fuse LiDAR odometry or SLAM with high-end INS and wheel odometry) as a baseline [37] or directly measure the start-to-end drift [38] for evaluation.

- *3D Structure*:

Chronically, it is a bit unusual to provide 3D-structure ground truth in SLAM datasets [12], [40], [41], although mapping is an integral and crucial part of research. One primary reason could be the hardware burden – high-accuracy 3D-scanners (as shown in Fig. 6) that provide dense scene model are hard to afford for many institutes, let alone individual researchers. For objects or small-scale scenes, a portable 3D-scanner powered by structured light or laser could be the best choice (e.g., GOM ATOS[14], Artec 3D[15], and Shining 3D[16]), providing with an accuracy of 10-um level. For large-scale scenes, the best solution is to use a professional surveying instrument (e.g., FARO, Leica[17], and Trimble[18]), with an mm-level accuracy in hundreds of meters. The scanners are able to cover very huge area by registering multiple stations, with each individual scan consuming several to tens of minutes. Moreover, sometimes the area could be too large, then if the pose data are accurate, it is also applicable to use a mobile mapping system (MMS, e.g., RIEGL[19], which fuses high-precision RTK, LiDAR, and IMU) or take the mapping results of LiDAR-based mapping/SLAM algorithms as 3D structure baseline [37].

## B. *Fundamental Characteristics Dimensions*

As reflected by the collection methodology, the characteristics of SLAM related datasets are mainly decided by 5 fundamental dimensions: mobile platform, sensor setup (calibrated perception sensors), target scene, data sequence, and ground truth. Therefore, these dimensions will also be the basis of datasets introduction, comparison, and analyses in the following sections. The descriptions of each dimension are given below.

[9] https://www.vicon.com/
[10] https://optitrack.com/
[11] https://www.motionanalysis.com/
[12] https://www.faro.com/
[13] https://www.hexagonmi.com/products/laser-tracker-systems
[14] https://www.gom.com/metrology-systems/atos.html
[15] https://www.artec3d.com/
[16] https://www.shining3d.com/
[17] https://leica-geosystems.com/products/laser-scanners
[18] https://www.trimble.com/3d-laser-scanning/3d-scanners.aspx
[19] http://www.riegl.com/nc/products/mobile-scanning/

*1) Mobile Platform*

Mobile platform decides the motion pattern of a dataset. The general characteristics of several common platforms are listed in Table I, explaining the degree of freedom (DoF), speed, and motion properties. Such characteristics are identified to impact algorithm performance, for example, rapid motion (e.g., shake, sharp turn/rotation/shifting, and high-speed) can affect the data quality, such as camera motion blur (see Fig. 7-left), and LiDAR distortion (both 2D/3D LiDARs are designed to sweep the laser emitter to scan massive points asynchronously, which are packed into a single frame but with indeed different coordinates in the motion, badly messing the result, see Fig. 7-right). Besides, pure rotation will cause fundamental matrix degeneration in monocular case, and fast speed may cause narrow overlap between adjacent frames. As shown in the table, generally 6-DoF platforms (handheld, backpack, UAV) are more difficult to algorithms.

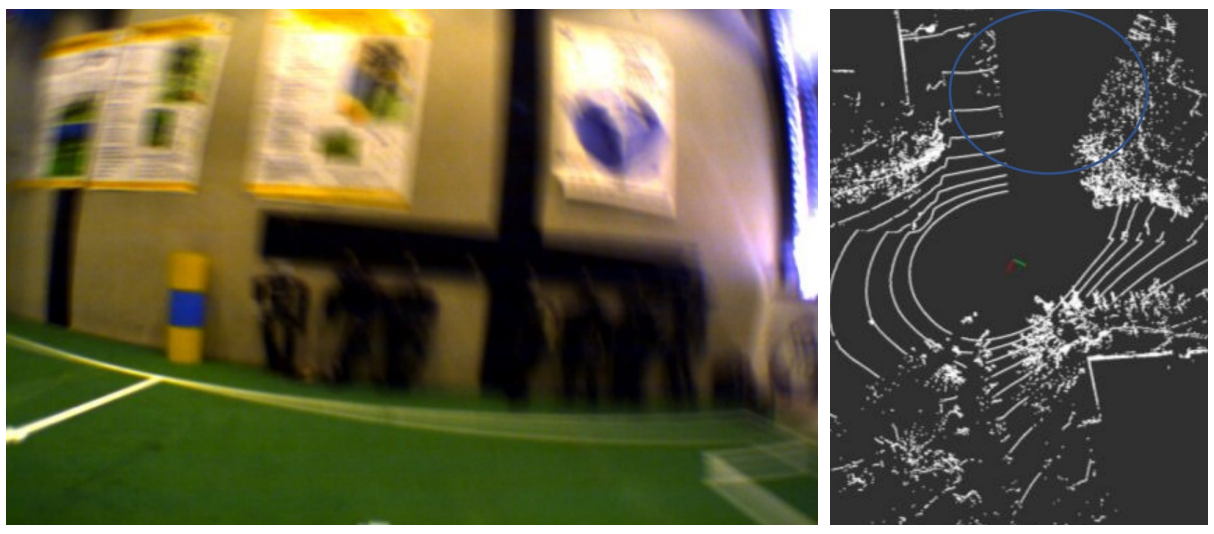

Fig. 7. Camera motion blur under walking shake (left), and LiDAR motion distortion under rapid rotation (right) [42], [43].

TABLE I
THE CHARACTERISTICS OF DIFFERENT MOBILE PLATFORMS

| Platform | DOF | Speed | Motion Properties |
|---|---|---|---|
| Handheld | 6 | ≈ 1m/s | Frequent shake, sharp turn, rapid rotation, sharp shifting |
| Backpack | 6 | ≈ 1m/s | Frequent shake |
| Automobile | 3 | 10-25m/s | Smooth forward/turn/shifting, constant speed |
| Wheeled robot | 3 | 1-5m/s | Smooth forward/turn/shifting, constant speed |
| UAV | 6 | 1-15m/s | Turn, rotation, sharp shifting |

*2) Sensor Setup*

Sensor setup refers to the number and modalities of calibrated perception sensors, thus decides the supported algorithm types of a dataset. As illustrated in the methodology, we can assume that all the perception sensors have been calibrated well, thus they can be used in any combinations. For example, if the dataset is with the sensor setup of binocular stereo cameras, a LiDAR, and an IMU, then it can support algorithms based on stereo vision, vision-inertial fusion (VI), vision-LiDAR fusion (VL), vision-LiDAR-inertial fusion (VIL), etc. When building or selecting a dataset, it is the primary factor to be considered.

*3) Target Scene*

Target scene involves the type, condition, and scale of the expected test environment. The type of the scene decides the data appearance and structure, which are exactly what the algorithms rely on to estimate motions, thus directly affect the performance. Imagine that if a camera is moving in front of a pure-white wall (namely weak texture), then it is impossible for a vision-based algorithm to extract salient features to track the motion. Besides, if a LiDAR-based method is working in a long corridor (namely repetitive structure), then it is quite tough to get match between LiDAR scans. Likewise, scene type can also drag in some other challenging elements include repetitive texture (grassland, sand, etc.), weak structure (open field, farmland, etc.), and interference feature and structure (reflection of water and glass, etc.).

Scene condition mainly include illumination, dynamic objects, weather, season, and time. These elements all have impacts on algorithms performance, for example, under bad illumination or at dark night a camera cannot acquire good images.

Scene scale is also important, as it indirectly affect the volume and variety of data sequences. It is largely determined by scene type, for example, an urban or campus scene can usually cover a large scale, while a room scene usually has a small area.

*4) Data Sequence*

Data sequence involves the quantity of sequences, the length of paths, and the variety of collection routes. Usually, datasets with greater quantity, longer trajectories, and various routes have more complete coverage of the scene, thus have more chance to inspect the weaknesses of the algorithms.

*5) Ground Truth*

Ground truth involves the types and quality of the reference data, it decides the evaluable tasks of a dataset. The previous methodology part has given a comprehensive introduction on the existing ground truth techniques, while they are quite different in applicable scenario, requirements, and accuracy. Here a characteristics comparison among them is given in Table II. It should be noted that, although with different accuracy, mostly the ground truth data are already quite enough for SLAM-related evaluation.

TABLE II
THE COMPARISON OF THE CHARACTERISTICS AMONG SOME COMMONLY USED GROUND TRUTH TECHNIQUES

| Technique | GT | Scenario | Accuracy |
|---|---|---|---|
| MoCap | 6-DoF pose | Indoor (small scale) | High: sub- mm/deg |
| RTK-GNSS (&INS) | 6-DoF pose | Outdoor | High: cm level (open sky) Low: denied area |
| Laser Tracker | 3D position | In/outdoor | High: sub- mm (occlusion free) |
| SLAM-based algorithm | 6-DoF pose | In/outdoor | Low/Medium |
| Marker, WiFi, … | 6-DoF pose/ 3D position | In/outdoor (middle scale) | Medium: cm-dm level |
| Start-to-end drift | 0 loop-drift | In/outdoor | Low |
| Large-scale 3D scanner | 3D structure | Scene/object | High: mm level |
| Small-scale 3D scanner | 3D structure | Object | High: sub-mm level |
| SLAM-based algorithm | 3D structure | Scene/object | Low/Medium |

## III. SLAM RELATED TASKS TAXONOMY AND DATASETS CATEGORIZATION

This section provides a clear SLAM related tasks taxonomy and datasets categorization, in the hope of resolving the two-way selection difficulty between algorithm researchers and datasets providers.

The proposed taxonomy of SLAM related tasks takes into account 2 principles: sensor-oriented, and functionality-oriented.

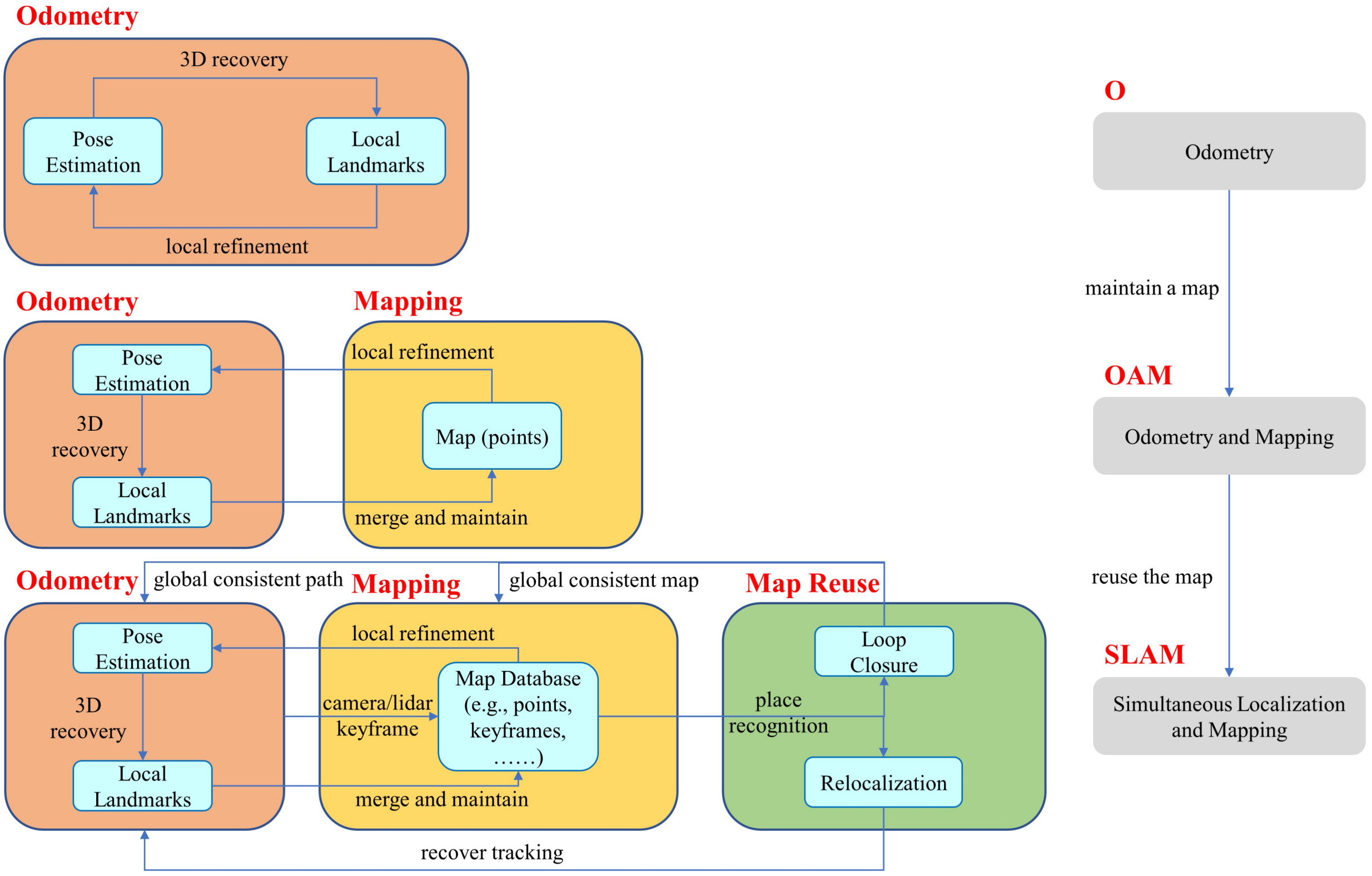

Fig. 8. The technical architecture and hierarchical relationship among odometry, mapping, and SLAM.

TABLE III
THE OVERALL TAXONOMY OF SLAM RELATED TASKS

| Sensor Setup | | | | | Functionality |
|---|---|---|---|---|---|
| **Based Sensor** | **Sub-sensor-modality** | | **Aiding Sensor** | | |
| Vision-based (V) | Monocular, Stereo, Fisheye, Omnidirectional, Thermal, Event, … | | IMU (I)/ | | Odometry (O)/ |
| LiDAR-based (L) | 2D LiDAR, 3D LiDAR | | GNSS/ | | Odometry and Mapping (OAM)/ |
| Vision-LiDAR-based (VL) | Combination of the above | ⊕ | Wheel Encoder/ | ⊕ | Simultaneous Localization and Mapping (SLAM)/ |
| RGB-D-based | N/A | | …… | | 3D Reconstruction (3DR) |
| Image-based | N/A | | N/A | | Structure from Motion (SfM)/ Multiple View Stereo (MVS) |

Oriented by sensor setup, there are mainly five branches for SLAM related tasks: LiDAR-based, vision-based, vision-LiDAR fusion, RGB-D-based, and image-based. Benefiting from the high-accuracy ranging of laser sensors, LiDAR-based methods (2D [44] and 3D [45]) are widely employed, acting as a mature solution. However, the price of LiDAR is too high, with multi-beam editions (3D) costing at least thousands of dollars [46]. As the visual sensors become ubiquitous, along with the improvements of computing hardware, vision-based methods have greatly attracted the efforts of SLAM community [47], [48], [49]. More specifically, due to the diversity of visual modalities, there has emerged many sub-classes, including monocular, stereo, fisheye, event-based [50] methods, etc., suitable for different application scenarios. Although not as robust nor accurate as LiDAR-based methods [45], [51], [52], still vision methods are proved to have huge potentials and are steadily on its way forward [53]. In order to combine the advantages of the both modalities, vision-LiDAR fusion has become a hot topic that attracts significant attention [54]. RGB-D is a novel modality that also combines the imaging and ranging sensor. It collects color images along with a depth channel which make it much easier to perform 3D perception. Since the Microsoft Kinect [22] was first introduced, RGB-D sensors have been widely adopted in SLAM research due to the low-cost and complementary nature [55], [56]. Additionally, there is another branch named image-based which is popular in 3D computer vision. Compared with vision-based methods, it does not require specifically arranged sensor setup or sequential images, photos that collected from the Internet can also be used to recover the 3D model of a scene or an object. For the other modalities, such as IMU, GNSS, and wheel odometry, they can provide a priori and be fused into optimization thus enhance the accuracy and robustness of motion estimation. Since they are adopted mainly to play aiding roles, the related fusions are categorized into the above five branches.

In the context of the above sensor setups, oriented by functionality, there are two main branches of SLAM related problems: mobile localization and mapping, and 3D reconstruction.

Within the range of mobile localization and mapping, there are 3 specific tasks: odometry, mapping (here we refer to a real-time or quasi real-time sequential manner), and SLAM. The 3 terms are with a hierarchical relationship at technical level, as shown in Fig. 8. Odometry means to track the position and orientation of an agent incrementally over time [57]. The abbreviation of such system is usually named as "O", such as VO, VIO, and LO, which refers to Visual Odometry, Visual-Inertial Odometry, and LiDAR Odometry respectively. It was proposed to serve as an alternative or supplement to wheel odometry due to the better performance especially in uneven or other adverse terrains [58], [59], [60]. In the strict sense, besides tracking the path, odometry also recovers 3D points, whereas these landmarks are only used for local optimization rather than a map. Mapping means to build a map of the environment in a particular format through mobile collection. Normally, mapping works along with odometry, thus named as odometry and mapping (OAM) (we avoid using the term mobile mapping as it usually refers to a more professional procedure in surveying and mapping domain [61]), since it relies on pose estimation to recover 3D points and merge together scene structures. As for SLAM, intuitively, it has both functions of odometry and mapping, but it is not a simple combination of the two. SLAM means to keep a track of an agent's location while at the same time building a global consistent and reusable map [57], [62]. This implies two core functions of SLAM: loop closure and relocalization, which rely on the map database to query the previous visited places. Loop closure means to add new constraints around revisited places to perform global optimization, thus eliminate accumulated drift of the path and ensure a global consistent map. Relocalization means to recover positioning within the prebuilt map after the robot lose its track. As can be seen, the "SLAM map" is not only a model, but also a database for query, which supports the map reuse capability that draws a line between SLAM and the other two. It should be noted that, the map does not have to be dense or good looking, the point is whether the robot can recognize. After all, sparse map points can also store features and descriptors [63], [64].

3D reconstruction means to recover a high-quality 3D map or model of a scene or an object from sequential data or data of different locations and perspectives. The general methodology is similar with mobile localization and mapping, but it does not require sequential data and real-time running. It can establish as many as possible constraints among any co-visible data frames and does much more optimization, thus no doubt achieve better accuracy and density. It can be achieved with many modalities of sensors (e.g., camera, LiDAR, and RGB-D), but here we want to emphasize image-based approach a bit, which does not require sequential data or a fixed sensor. There are mainly two specific tasks in image-based field: structure from motion (SfM) and multiple view stereo (MVS). SfM is a process of recovering 3D structure from 2D images of different locations and views [65], by estimating camera poses among any co-visible views. Usually, the output models are composed of sparse point cloud. MVS can be regarded as a generalization of the two-view stereo, it assumes known camera poses of each view (just as the extrinsic parameter between binocular cameras) to recover dense 3D models. Since SfM already estimates the camera poses beforehand, MVS can directly serve as a post process of SfM to build an accurate and dense model [66].

Overall, comprehensively considering the above principles and the consensus of this field, we propose the taxonomy as the combination of the based sensor, aiding sensor, and functionality, as shown in Table III. For example, if an algorithm is based on LiDAR and the functionality is Odometry and Mapping (OAM), then the task will be categorized as LOAM (this coincides with the name of the algorithm proposed by Zhang et al. [45]); if the algorithm is based on Vision (V), aided with an IMU (I), and the functionality is Odometry (O), then the task will be categorized as VIO. It should be noted that, within the aiding sensors, usually GNSS and wheel odometry are used for generating ground truth, but to fuse them into SLAM process is also doable and effective.

As for datasets categorization, logically, to match the datasets with corresponding algorithm types, we mainly follow the sensor setup as the basis. Although some methods will require specific datasets (such as event-based VO/VSLAM [50]), considering that typically a dataset is not only designed for one task, to maintain the compatibility and selection coverage, we avoid categorizing the datasets too fine, but instead resulting in 4 broad categories: 1. Vision-based mobile localization and mapping and 3D reconstruction datasets, 2. LiDAR-related mobile localization and mapping and 3D reconstruction datasets, 3. RGB-D-based mobile localization and mapping and 3D reconstruction datasets, and 4. Image-based 3D reconstruction datasets. This criterion coincides with the sensor setup of the proposed SLAM tasks taxonomy, whereas the only difference is that the datasets suitable for LiDAR-based and vision-LiDAR-based algorithms are merged together into LiDAR-related category. Because, from the dataset point of view, if the platform is equipped with LiDAR, then it can also accommodate cameras. It should be noted that, RGB-D datasets and LiDAR-related datasets that contain camera data can also support vision-based methods. For example, the best-known KITTI dataset [14] and TUM RGB-D dataset [12] are usually used for benchmarking VO and VSLAM algorithms. Moreover, some datasets with IMU data [67] and 6-DoF motions could be originally designed for IMU fusion algorithms, thus are typically with higher difficulty. To avoid possible confusions, we will further provide a considerate dataset selection guideline in Section V.

## IV. State-of-the-art Datasets

This section gives a structured and detailed introduction of 16 state-of-the-art datasets by the above 4 broad categories, which we believe can well represent the whole datasets entirety, to help understand current datasets and also to serve as a good lesson for future releases. Many deep insights that even not covered by the original papers are given, making the enumeration worthy. Note that, limited by space, we avoid introducing more cases minutely, which on the other hand will also not make more sense due to the homogeneous information. Hence, such selection does not mean any recommendation bias, and it is exactly one main goal of this paper to remove the biased usage. To achieve clear descriptions, the datasets are introduced in a structured way, mainly attending to general information, platform and sensor setup, scenes and sequences, ground truth, and highlights. Although introduced individually, we do follow a chronological order, and emphasize the features and progresses compared with previous works. For a holistic comparison and analysis, we will arrange a concise but comprehensive overview covering all the datasets in Section V, and give analyzation and discussion in Section VII.

### *A. Vision-based Mobile Localization And Mapping And 3D Reconstruction Category*

#### *1) EuRoC MAV* [15]

The EuRoC MAV dataset was collected under the context of the European Robotics Challenge (EuRoC), in particular, for the Micro Aerial Vehicle competitions of visual-inertial algorithms. Since the publication in 2016, it has been tested by a great many teams and cited by massive literature, becoming one of the most widely used datasets in SLAM scope.

TABLE IV
THE PROPERTIES OF THE SENSORS AND GROUND TRUTH INSTRUMENTS

| Sensor/Instrument | Type | Rate | Characteristics |
|---|---|---|---|
| Camera | 2 × MT9V034 | 20 Hz | WVGA, monochrome, global shutter |
| IMU | ADIS16448 | 200 Hz | MEMS, 3D accelerometer, 3D gyroscope |
| Laser Tracker | Leica MS50 | 20 Hz | 1mm accuracy |
| Motion Capture | Vicon | 100 Hz | sub-(mm/deg) accuracy |
| 3D Scanner | Leica MS50 | 30 kHz | 3mm accuracy |

• *Platform and sensor setup*: The dataset was collected using an MAV, logically designed to provide 6-DoF motions with slow to high speed. Two grayscale cameras (stereo setup) and an IMU were equipped and with strictly spatial-temporal alignment (the properties are listed in Table IV), thus ideal for vision (mono and stereo) and especially IMU fusion algorithms.

• *Scenes and sequences*: The dataset consists of two portions: machine hall and room, with 5 and 6 sequences respectively. Varying in terms of texture, motion, and illumination conditions, the datasets were recorded in 3 difficulty levels: easy, medium, and difficult. This has been proved to be challenging for many algorithms [49], [68], [69], enabling researchers to locate the weakness and enhance their algorithms effectively.

• *Ground truth*: The dataset has reliable ground truth data of both positioning and 3D mapping, thus support both kinds of evaluation. For the first batch collected in the machine hall, the 3D position ground truth was provided by laser tracker. For the second batch in the room, both 6D pose and 3D map ground truth were provided, respectively by motion capture system and laser scanner. All the ground truth is with extreme quality, as shown in Table IV from the instrument properties.

• *Highlights*: Vision-IMU setup, multiple complexity-levels, highly accurate ground truth, 3D map ground truth.

#### *2) TUM MonoVO* [38]

TUM MonoVO dataset, released in 2016, was specifically for testing long-range tracking accuracy of O/SLAM, complementing previous datasets with larger scale. Markedly, all the images have been photometrically calibrated.

• *Platform and sensor setup*: The dataset was collected through a handheld carrier, which makes the motion pattern challenging. Two grayscale cameras were used, but merely with wide and narrow-angle lenses rather than forming a stereo pair, so can only be used for monocular algorithms. The cameras have up to 60Hz frame rate, which are far above normal to enable handling quick and unsmooth motions while designing algorithms.

• *Scenes and sequences*: The dataset provides 50 sequences in a variety of scenes from indoor to outdoor. The indoor scenes were mainly recorded in a school building, covering office, corridor, large hall, and so on. The outdoor scenes were mainly recorded in a campus area, including building, square, parking lot, and the like. Many sequences have extremely long distances, which are expected by researchers to perform long-range tests.

• *Ground truth*: Due to the massive collections in indoor and outdoor integration environments, it was almost impossible to measure ground truth using GNSS/INS or motion capture system. So this dataset was designed to record each sequence starting and ending at the same position, and generated a set of "ground truth" poses at the start and end segments, allowing to evaluate the tracking accuracy based on the loop drift. Note that, you have no chance to inspect the exact locations where the errors occur, hence not recommended if you target frame-by-frame accuracy. And you must disable the "loop closure" function for SLAM usages, otherwise the zero drift will invalid the evaluation.

• *Highlights*: Handheld motion pattern, diverse scenes various sequences, photometrically calibration, in-outdoor integration.

#### *3) UZH-FPV Drone Racing* [70]

The UZH-FPV Drone Racing dataset, published in 2019, is officially used in IROS 2020 Drone Racing Competition. Among the existing real-world datasets, it contains the most aggressive motions, complementing previous 6-DoF datasets further.

• *Platform and sensor setup*: The dataset was collected using a quadrotor controlled by an expert pilot to make extreme speeds, accelerations, and fast rotations, proved to be far beyond the capabilities of existing algorithms [71]. It equipped a grayscale event camera, and a grayscale binocular stereo camera, both with hardware synchronized IMU (properties are listed in Table V), thus can support quite well vision-based (mono and stereo), IMU fusion, and event-based methods.

TABLE V
THE PROPERTIES OF THE SENSORS AND GROUND TRUTH INSTRUMENT

| Sensor/Instrument | Type | Rate | Characteristics |
|---|---|---|---|
| Event Camera | mDAVIS | 50 Hz | 346 × 260, grayscale + events, 120° FOV |
| IMU #1 | mDAVIS built-in | 500Hz | 3D accelerometer, 3D gyroscope |
| Stereo Camera (binocular) | OV7251 | 120 Hz | 640 × 480, grayscale, global shutter, 186° FOV |
| IMU #2 | Snapdragon built-in | 1000Hz | 3D accelerometer, 3D gyroscope, 3D magnetometer |
| Laser Tracker | Leica MS60 | 20 Hz | 1mm accuracy |

• *Scenes and sequences*: The dataset was collected in an indoor airplane hangar and an outdoor grassland edged by trees, forming a total of 27 sequences. It has a 340.1/923.5 (m) max distance and a 12.8/23.4 (m/s) top speed for indoor/outdoor scene (both the top speeds are larger than any existing sequences).

• *Ground truth*: The ground truth were generated by tracking the 3D position using the Leica MS60 total station, with below 1mm accuracy. However, due to the aggressive flights, laser lost tracking could always occur causing trajectory broken, thus only complete sequences were selected for publication.

• *Highlights*: Aggressive motion, highly accurate ground truth, event data.

4) *OpenLORIS-Scene* [41]

OpenLORIS-Scene, published in 2019, is the official dataset of IROS 2019 Lifelong SLAM Challenge. Compared with previous datasets, it is a real-world collection in dynamic and daily-changing scenarios specialized for long-term robot navigation.

• *Platform and sensor setup*: The dataset was collected by wheeled robots (with odometry) with an RGB-D camera and a color binocular stereo camera (both with hardware synchronized IMU), and a 2D/3D LiDAR (all properties are listed in Table VI). It well supports monocular, binocular, RGB-D, and IMU fusion methods, but not recommended for LiDAR-based, since LiDAR is already used for generating ground truth.

TABLE VI
THE PROPERTIES OF THE SENSORS AND GROUND TRUTH INSTRUMENTS

| Sensor/Instrument | Type | Rate | Characteristics |
|---|---|---|---|
| RGB-D Camera | RealSense D435i | 30 Hz | 848 × 480,<br>RGB + Depth,<br>global shutter,<br>69° × 42°FOV,<br>6-axis IMU |
| Stereo Camera (binocular) | RealSense T265 | 30 Hz | 640 × 480,<br>RGB,<br>global shutter,<br>186°FOV,<br>6-axis IMU |
| LiDAR #1 | Hokuyo UTM-30LX | 40 Hz | 2D LiDAR,<br>270° Horizontal,<br>30m range,<br>5cm accuracy, |
| LiDAR #2 | RoboSense RS-LiDAR-16 | 20 Hz | 16-beam LiDAR,<br>360° H × 30° V,<br>150m range,<br>2cm accuracy |
| Motion Capture | OptiTrack | 240 Hz | sub-(mm/deg) accuracy |

• *Scenes and sequences*: The dataset was collected in 5 scenes: office, corridor, home, café, and market, which are quite typical application scenarios for service robots but rarely appeared in previous datasets. Thanks to the real-life record, many dynamic people and objects were brought inside, fulfilling the pressing expectation of such data. It also targets long-term SLAM, thus many sequences of each same scene but at different time slots were captured, which incorporates ever-change of illumination, viewpoints, and those caused by human activities (see Fig. 9).

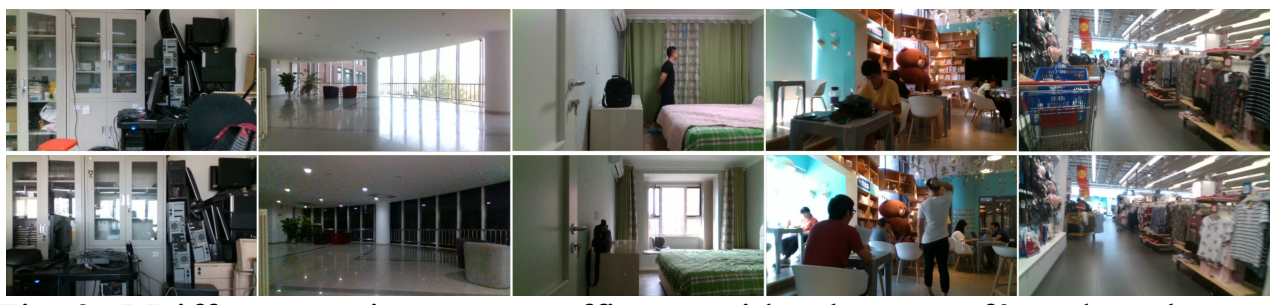

Fig. 9. 5 Different environments: office, corridor, home, café, and market from left to right collected at different time (top and bottom row) [41].

• *Ground truth*: For office scene, the ground truth was provided by Motion Capture System at 240 Hz, ideal for frame-by-frame evaluation. For the other scenes, the ground truth was generated by LiDAR SLAM and partly accompanied with sensor fusion, reported to have cm-level accuracy, which are not that strict but enough for vision methods evaluation.

• *Highlights*: Service robot platform, multi-modal sensors, typical scenes, ever-changing environments, long-term data.

5) *TartanAir* [72]

TartanAir, published in 2020, is the official dataset of CVPR 2020 Visual SLAM Challenge. It is a synthetic dataset collected via simulation, thus accomplish to design many challenging effects, with the mission of pushing the limits of visual SLAM.

• *Platform and sensor setup*: TartanAir is a simulated dataset, thus there is no practical platform or sensor setup. However, it created random motion patterns with high complexity, which are good for testing whereas not realistic as real-world motions. Also, multi-modal sensors are simulated: two color cameras (stereo setup), an IMU, a depth camera, and a 3D LiDAR, supporting many modals of SLAM related tasks, whereas such data are quite perfect, which fail to mimic the real-world camera blur or motion distortion of LiDAR, implying the weakness of such manner.

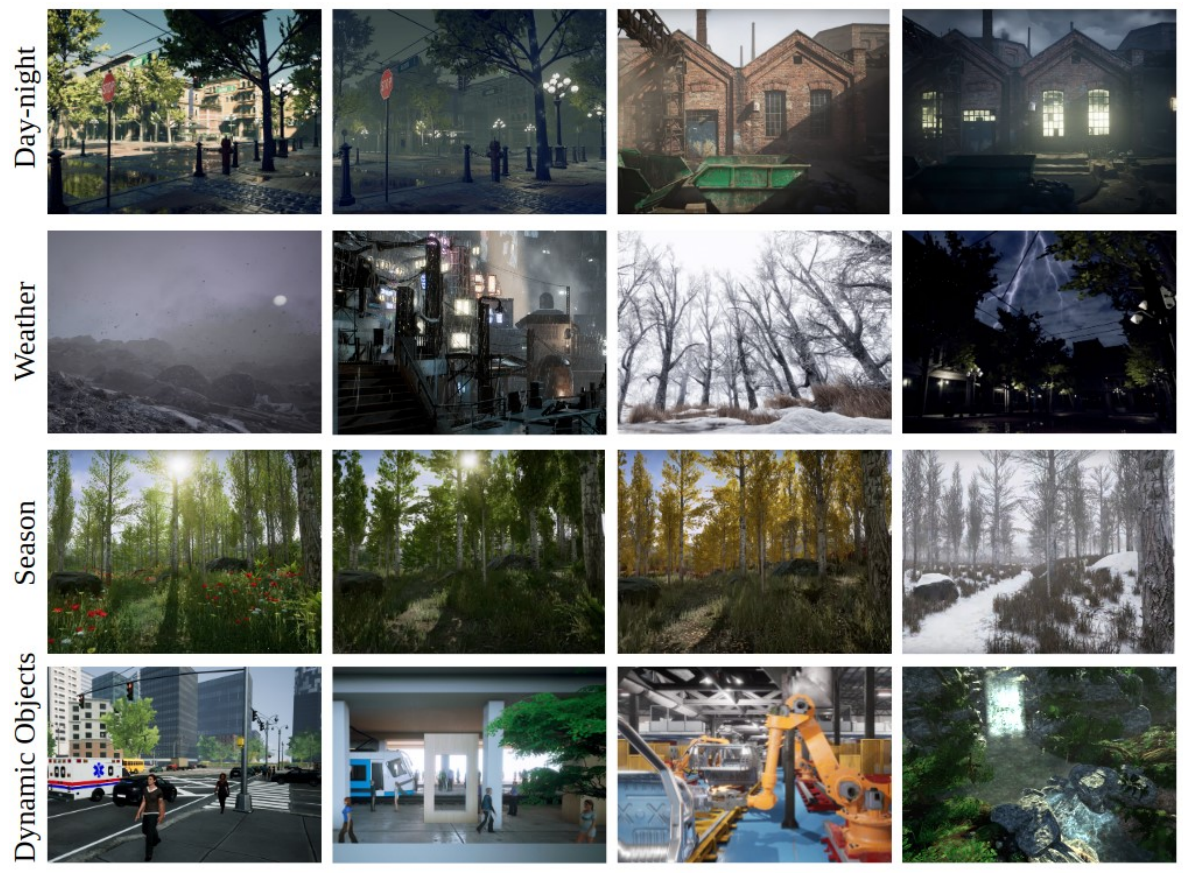

Fig. 10. Some representative images of different categories: Urban, Rural, Nature, Domestic, Public, and Scifi from left column to right [72].

• *Scenes and sequences*: TartanAir has simulated diverse scenes (30 in 6 categories): Urban, Rural, Nature, Domestic, Public, and Scifi. As shown in Fig. 10, the scenes were designed with many challenging effects, such as light changing, bad illumination, adverse weather, time and season effects, and dynamic objects. The dataset has a total of 1037 long sequences, resulting in more than 1M frames, which are much larger than existing datasets. However, although seem plentiful the images, when inspecting at a full resolution, still you will find the quality far from realistic.

• *Ground truth*: Both camera poses and 3D map ground truth are provided. Notably, they are absolutely accurate thanks to the simulation manner. Besides, some other labels are also provided, such as semantic segmentation, stereo disparity, depth map, and optical flow.

• *Highlights*: Diverse motion patterns, diverse scenes, wealthy sequences, challenging effects, perfect ground truth.

### B. *LiDAR-related Mobile Localization And Mapping And 3D Reconstruction Category*

1) *KITTI Odometry* [14]

The KITTI Odometry dataset, released by Karlsruhe Institute of Technology (KIT) and Toyota Technological Institute in 2012, is the most popular dataset in SLAM domain. Except for the high-quality dataset itself, it is also largely because of the famous public benchmark.

• *Platform and sensor setup*: KITTI was collected using a car with binocular stereo color and monochrome cameras (both pairs triggered via hardware and with a long baseline), a LiDAR (synchronized per sweep with camera frames), and an RTK-GPS&INS device (for ground truth). All the sensors are top-level (the properties are listed in Table VII) and offer high-quality data, supporting vision-based (mono and stereo) and LiDAR related O/SLAM, whereas conversely the smooth motion, clear images, and dense point clouds do make the algorithms easy to handle, which may be insufficient for the demonstration of normal cases.

• *Scenes and sequences*: The dataset was collected around a mid-size city of Karlsruhe, in rural areas and on highways. It contains a huge volume of raw data, while considering the long trajectories, varying speeds, and reliable GPS signal, a total of 39.2km's driving distance were selected, forming 22 sequences with frequent loop closures, ideal for SLAM validation purpose. However, although collected in real life, still KITTI is generally known to be easy: good lighting and weather, sufficient texture and structure, and static scene in most sequences, leading to an impressive result of 0.53% error in translation [73]. It is worth noting that, the 21th and 22th sequences captured on the highway are highly dynamic (see Fig. 11), which could be challenging for motion estimation algorithms.

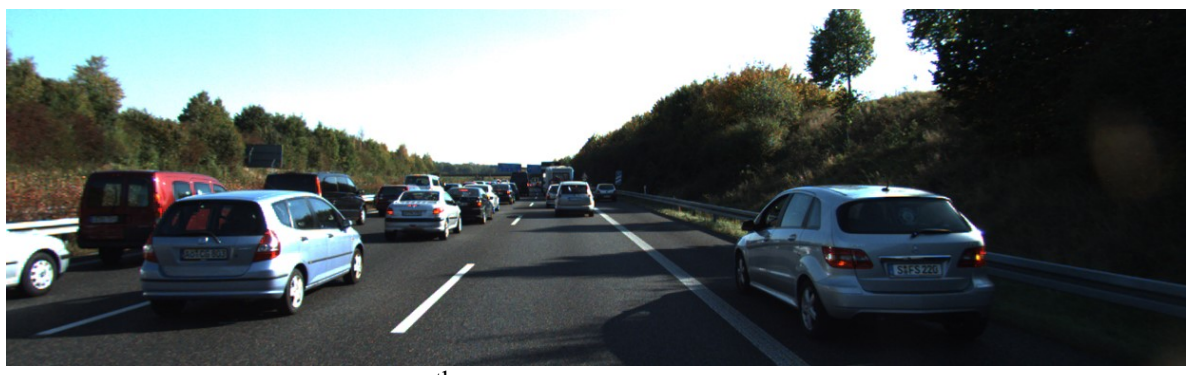

Fig. 11. A frame from the 21th sequence that crowded with moving cars.

TABLE VII
THE PROPERTIES OF THE SENSORS AND GROUND TRUTH DEVICES

| Sensor/Devices | Type | Rate | Characteristics |
|---|---|---|---|
| Camera (monochrome) | 2 × PointGrey Flea2 FL2-14S3M-C | 10 Hz | 1392 × 512, 8-bit gray, global shutter, 90° × 35° FOV |
| Camera (color) | 2 × PointGrey Flea2 FL2-14S3C-C | 10 Hz | 1392 × 512, 8-bit RGB, global shutter, 90° × 35° FOV |
| LiDAR | Velodyne HDL-64E | 10 Hz | 64-beam LiDAR, 360° H × 26.8° V, 120m range, 2cm accuracy |
| RTK-GPS&INS | OXTS RT3003 | 100 Hz | cm/sub-deg accuracy |

• *Ground truth*: A high-end RTK-GPS&INS device was used to generate pose ground truth. Trajectories with good GPS signal were selected to ensure cm level accuracy. But note that, due to the GPS nature of absolute measurement, the ground truth was not smooth locally, which will cause biased evaluation results if studying very small sub-sequence. An intuitive understanding is you cannot measure accurately the thickness of a single piece of paper, but are able to measure 100 pages together. So, in 2013, KITTI changed the minimum evaluated length from 5 to 100 and counting longer sequences, leading to a better indication on performance. This also gives us a lesson: if your technique cannot give accurate ground truth, focus longer segments for evaluation (to study start-to-end drift is an extreme version of this concept, such as TUM MonoVO).

• *Highlights*: High quality vision-LiDAR data, accurate ground truth, wealthy sequences, loop closures, public benchmark.

2) *Oxford RobotCar* [40]

The Oxford RobotCar dataset, presented in 2016, was recorded by traversing a same route twice a week in around one year, resulting in over 1000km's trajectories. Various time slots, weather, and scene change were covered inside, complementing previous datasets (such as KITTI) greatly with long term and challenging data.

• *Platform and sensor setup*: There are 4 cameras (1 trinocular stereo and 3 monocular), two 2D-LiDARs, a 3D-LiDAR, and a set of RTK-GPS&INS device (for ground truth) equipped on a car, thus supports vision-based (mono and stereo) and LiDAR related O/SLAM. The sensor properties are listed in Table VIII.

TABLE VIII
THE PROPERTIES OF THE SENSORS AND GROUND TRUTH DEVICES

| Sensor/Devices | Type | Rate | Characteristics |
|---|---|---|---|
| Stereo Camera (trinocular) | FLIR Bumblebee XB3 | 16 Hz | 1280 × 960, 8-bit RGB, global shutter, 66° HFOV, 12/24cm baseline |
| Camera | 3 × FLIR GS2-FW-14S5C-C | 12 Hz | 1024 × 1024, 8-bit RGB, global shutter, 180° HFOV |
| 2D LiDAR | SICK LMS-151 | 50 Hz | 270° Horizontal, 50m range, 3cm accuracy |
| 3D LiDAR | SICK LD-MRS (4 planes) | 12.5 Hz | 85° H × 3.2° V, 50m range, 3cm accuracy |
| RTK-GPS&INS | NovAtel SPAN-CPT | 10 Hz | cm/sub-deg accuracy |

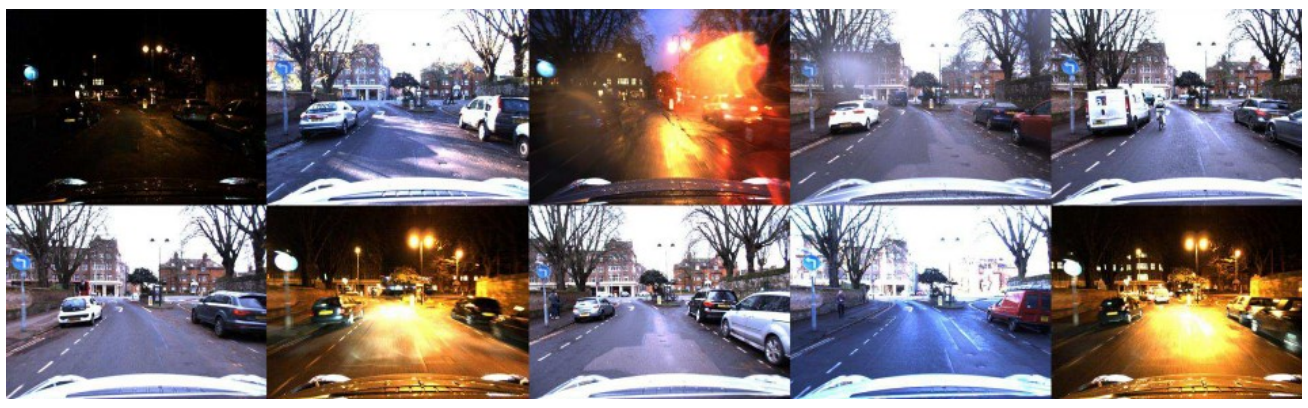

Fig. 12. The variation of conditions at the very same locations caused by time slots, weather, season, and human activities [74].

• *Scenes and sequences*: Due to the long-term frequent collection, a wide variety of scene appearances and structures caused by illumination change, diverse weather, dynamic objects, seasonal effects, and construction works were collected, covering inside pedestrian, cyclist, and vehicle traffic, light and heavy rain, direct sun, snow, and down, dusk and night. Compared with KITTI, RobotCar is more comprehensive at the level of mimicking the real world, thus will be more persuasive if algorithms can handle such variation and challenge with good scores. Several example images are shown in Fig. 12 to give an intuitive feel.

• *Ground truth*: Separated from the dataset, the ground truth of RobotCar was released in 2020 [75]. After well selected from the whole dataset, a sample of 72 traversals were provided 6D-pose ground truth at 10Hz by post-processing the high-end GNSS/INS with RTK correction, resulting in an cm/sub-deg accuracy.

• *Highlights*: Long-term dataset, long trajectory, vision-LiDAR setup, repeatedly traverse, various and challenging conditions.

3) *Complex Urban* [37]

The Complex Urban dataset, released in 2019, was recorded in diverse and complex urban environments suffering from denied or inaccurate GNSS, complementing KITTI and RobotCar with a totally different urban style.

- *Platform and sensor setup*: The dataset was collected using a car equipped with multi-modal sensors of different levels: two color cameras (binocular stereo), two 2D-LiDARs, two 3D-LiDARs, a consumer-level GPS, a VRS-RTK GPS (for ground truth), a 3-axis FOG (for ground truth), a consumer-level IMU, and two wheel encoders (for ground truth), supporting a wide range of SLAM related tasks based on vision (mono and stereo), LiDAR, and IMU/GNSS fusion. The sensors properties are listed in Table IX. But note that, due to the non-overlap, it does not support to project LiDAR points to the exact position on camera frames while building vision-LiDAR fusion algorithms.

TABLE IX
THE PROPERTIES OF THE SENSORS AND GROUND TRUTH DEVICES

| Sensor/Devices | Type | Rate | Characteristics |
|---|---|---|---|
| Camera | 2 × FLIR FL3-U3-20E4C-C | 10 Hz | 1280 × 560, 8-bit RGB, global shutter |
| 2D LiDAR | 2 × SICK LMS-511 | 100 Hz | 190° Horizontal, 80m range, 5cm accuracy |
| 3D LiDAR | 2 × Velodyne VLP-16 | 10 Hz | 16-beam LiDAR, 360° H × 30° V, 100m range, 3cm accuracy |
| GPS | U-Blox EVK-7P | 10 Hz | 2.5m accuracy, consumer-grade |
| FOG | KVH DSP-1760 | 1000 Hz | 3-axis, 0.05°/h bias |
| IMU | Xsens MTi-300 | 200 Hz | 9-axis AHRS, 10°/h bias |
| Encoder | 2 × RLS LM13 | 100 Hz | 4096 resolution |
| RTK-GPS | SOKKIA GRX2 | 1 Hz | H: 10mm, V: 15mm accuracy |

- *Scenes and sequences*: The dataset was collected in diverse and complex urban scenes such as metropolitan area, residential area, complex apartment, highway, tunnel, bridge, and campus. It covered inside many complex and high-rise buildings (shake the GNSS performance), multi-lane roads, and crowded moving objects, which are totally different with the European urban style of KITTI and the like, posing new challenge for algorithms. All the sequences have kilometers of trajectory length, enabling realistic and thorough tests. A sample frame and associated 3D point cloud are shown in Fig. 13, well illustrating "complex".

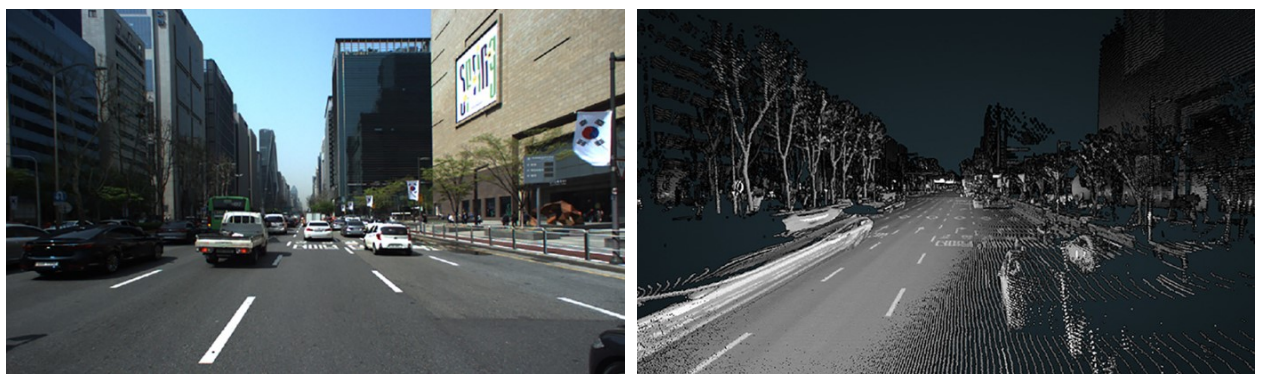

Fig. 13. A sample frame and associated 3D point cloud, containing dense moving objects and high-rise buildings, well illustrating "complex" [39].

- *Ground truth*: Due to the severe occlusion of high-rise buildings, bridges, and tunnels, it was impossible for VRS-GPS to calculate complete and accurate ground truth trajectories. To this end, the dataset has generated a baseline using an iSAM pose-graph SLAM framework [76], employing constraints from FOG, encoder, VRS-GPS, and ICP-based loop closure. See Table IX, the sensors and devices are in good properties, thus will not bias too much under such rigorous post-processing.
- *Highlights*: Multi-modal and multi-level sensors, complex urban environments, diverse scenes.

*4) Newer College* [77]

Newer College (the upgrade version of the 2009 New College dataset [78]), released in 2020, is the first LiDAR-related dataset collected in handheld motion pattern while at the same time with both trajectory and 3D mapping ground truth, complementing previous vehicular datasets significantly, thus is very precious.

- *Platform and sensor setup*: The dataset was collected using a handheld carrier, with 6-DoF motion containing vibration which can cause motion blur and LiDAR distortion. It was equipped an RGB-D sensor (only used as binocular stereo camera), and one 3D-LiDAR, both embedded with IMU, supporting vision-based (mono and stereo), LiDAR-related, and also IMU fusion SLAM related tasks. The sensors properties are shown in Table X.
- *Scenes and Sequences*: The dataset was collected in the New College of Oxford, with one sequence of 2.2km trajectory length. It was recorded at walking speeds (around 1 m/s) with several designed loop closures, thus can support the validation of SLAM and other place recognition algorithms quite well. The 3D map of the environment are shown in Fig. 14.

TABLE X
THE PROPERTIES OF THE SENSORS AND GROUND TRUTH INSTRUMENT

| Sensor/Instrument | Type | Rate | Characteristics |
|---|---|---|---|
| RGB-D Sensor (only used as binocular stereo camera) | RealSense D435i | 30 Hz | 2 × 848 × 480, RGB + Depth, global shutter, 69° × 42° FOV, 6-axis IMU @650hz |
| LiDAR | Ouster OS1-64 | 10 Hz | 64-beam LiDAR, 360° H × 45° V, 120m range, 5cm accuracy, 6-axis IMU @100hz |
| 3D scanner | Leica BLK360 | 360 kHz | 8mm accuracy |

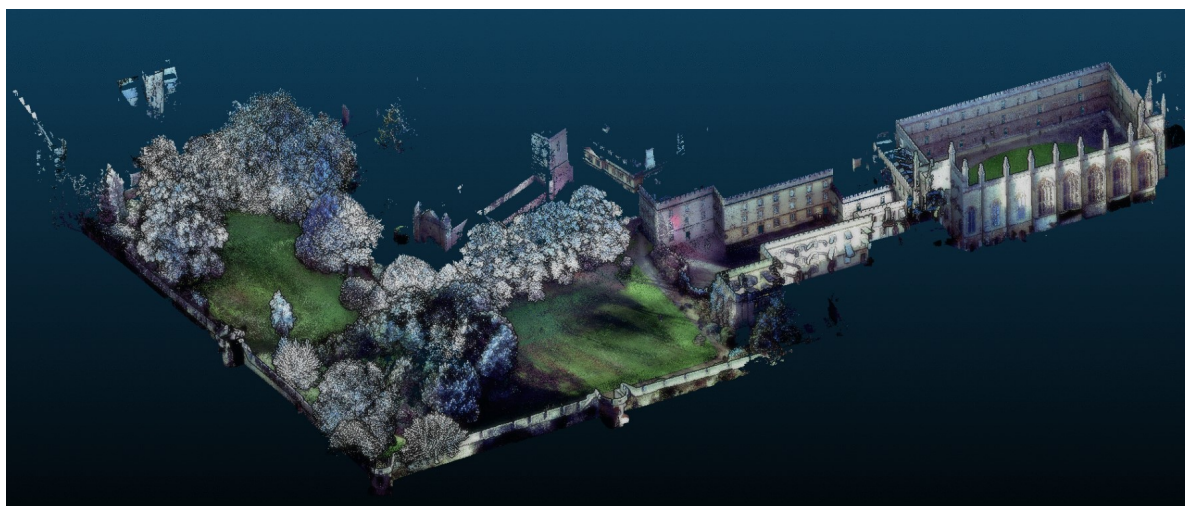

Fig. 14. The pre-build ground truth 3D map of the environment [77].

- *Ground truth*: The dataset provides both accurate pose and 3D map ground truth, thus support both localization and mapping evaluation, which is quite rare in previous datasets. As we have analyzed, it is not always accurate nor smooth to generate ground truth pose using GNSS, thus Newer College employed a novel method: first scan a dense ground truth 3D map of the scene with a survey-grade 3D laser scanner (as shown in Fig. 14), and then localize by registering the LiDAR data against the prebuild map (ICP was used here), thus a high-rate and accurate 6D-pose could be calculated. As listed in Table X, the 3D scanner has an <1cm accuracy while the LiDAR has cm-level, thus we can infer that the pose ground truth is also cm-level accurate. This is also a lesson for us to get ground truth pose when building middle-scale or indoor-outdoor integrated dataset.
- *Highlights*: Handheld LiDAR dataset, accurate pose and 3D map ground truth, long trajectory, loop closures.

### *C. RGB-D-based Mobile Localization And Mapping And 3D Reconstruction Category*

#### *1) TUM RGB-D* [12]

TUM RGB-D, released by Technical University of Munich in 2012, has become a top popular SLAM related dataset in the past decade. Though captured with RGB-D sensor, it is also widely used in pure vision-based methods. Remarkably, this dataset has proposed a set of evaluation criteria and indicators, becoming the consensus of localization evaluation till today (will be illustrated in Section VI).

• *Platform and sensor setup*: There are two recording platforms: handheld and robot, thus covering a wide range of motions. Both of them were equipped with an RGB-D sensor (Table XI lists the properties), supporting vision and RGB-D based O/SLAM.

TABLE XI
THE PROPERTIES OF THE SENSOR AND GROUND TRUTH INSTRUMENT

| Sensor/Instrument | Type | Rate | Characteristics |
|---|---|---|---|
| RGB-D Camera | Kinect v1 | 30 Hz | 640 × 480,<br>RGB + Depth,<br>rolling shutter,<br>$70^{\circ} \times 60^{\circ}$FOV,<br>3-axis accelerator |
| Motion Capture | MotionAnalysis Raptor-E | 100 Hz | sub-(mm/deg) accuracy |

• *Scenes and sequences*: The dataset was recorded in 2 different indoor scenes: an office (for handheld), and an industrial hall (for robot). A total of 39 sequences were captured, with and without loop closures, supporting SLAM validation.

• *Ground truth*: Both the scenes were fully covered by a motion capture system (Table XI lists the properties), providing high-rate and accurate 6D pose ground truth, supporting frame-by-frame evaluation. For ease of use, a set of automatic evaluation tools were provided. However, this dataset lacks 3D map ground truth, which could be the only glaring weakness.

• *Highlights*: Handheld and robot platforms, highly-accurate ground truth.

#### *2) ICL-NUIM* [79]

ICL-NUIM, released in 2014, is a synthetic dataset with both camera pose and 3D map ground truth, complementing the TUM RGB-D dataset with mapping evaluation.

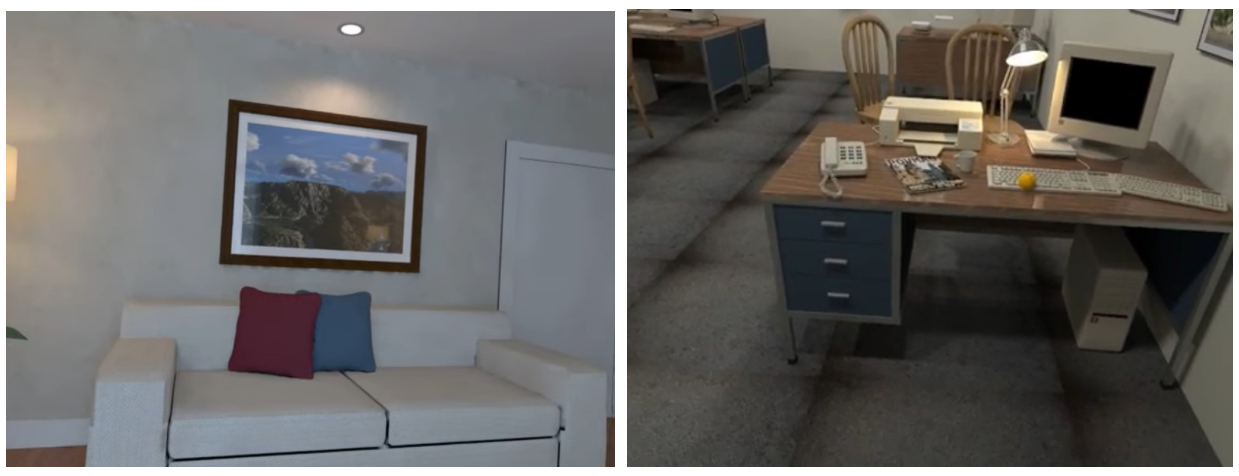

Fig. 15. Two example images: living room (left), and office room (right) [79].

• *Platform and sensor setup*: Different with many other synthetic datasets (e.g., TartanAir), ICL-NUIM duplicated a real-world handheld trajectory to make the motion pattern realistic. Then, an RGB-D sensor was simulated to capture the data within the 3D models in the POV-Ray software.

• *Scenes and sequences*: The dataset simulated 2 indoor scenes: a living room, and an office room. Each scene has 4 sequences, one of which designed with a small loop trajectory to support SLAM usage. Remarkably, noises were added to the RGB and depth images to mimic the real-world sensor noise. Such procedure made the data quite realistic, which is a good lesson for the datasets afterwards. As can be seen in Fig. 15, the rendered quality was even better than some later-released datasets, e.g., TartanAir. Whereas, it is a pity that both the scenes cover a relatively small area, with a maximal sequence length of only 11m.

• *Ground truth*: Since the collection already followed some pre-defined routes, they directly serve as perfect ground truth. The routes were obtained by running Kintinuous SLAM [80] from handheld data in a real living room, thus are quite realistic. For the office room scene, additionally the 3D model was provided as 3D mapping ground truth.

• *Highlights*: Realistic trajectory and data, accurate ground truth, 3D scene ground truth.

#### *3) InteriorNet* [81]

InteriorNet, presented in 2018, is a photo-realistic, large-scale, and highly variable synthetic dataset of interior scenes, complementing previous synthetic datasets (e.g., ICL-NUIM) markedly.

• *Platform and sensor setup*: To overcome the weakness of lacking trajectory realism in previous datasets, InteriorNet proposed to generate random trajectories first, and then augment them with a learned style model to mimic realistic motions. Compared with ICL-NUIM's duplicating real-world trajectories, this method is more flexible and efficient. As for sensor setup, except for RGB-D sensor, it also simulated IMU (noise available), stereo, events, and so on. Remarkably, the final frames were obtained by averaging between shutter open and close renderings, accomplishing to mimic motion blur, outperforming many other datasets.

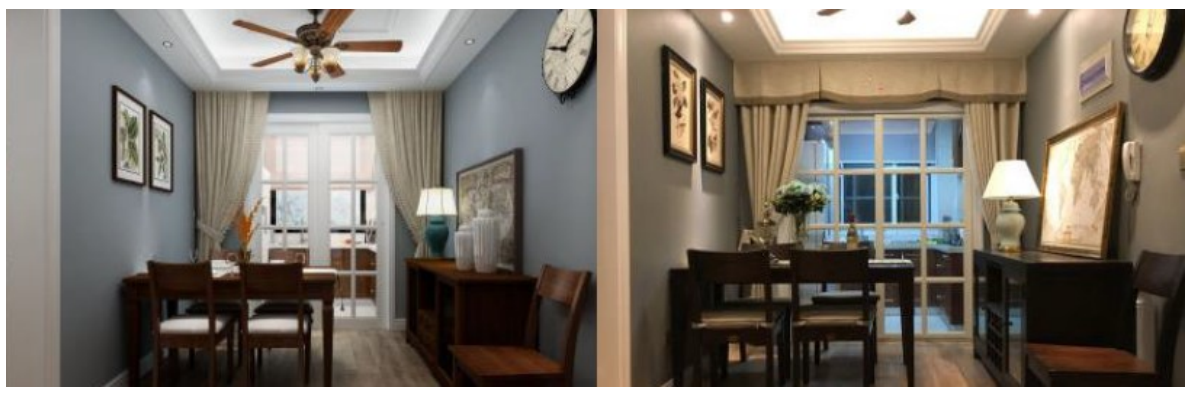

Fig. 16. The Rendering results (left) *vs*. real-world decorations (right) [81].

• *Scenes and sequences*: The dataset was rendered in various interior scenes and layouts collected from world-leading furniture manufacturers with fine geometric details and high-resolution texture, ranging from small studio to large 42-room apartment. A total of 15k sequences with 1k frames in each are provided, along with many challenging conditions, including various lighting, moving objects, and motion blur. The rendered images are very close to the real world, as depicted in Fig. 16.

• *Ground truth*: The 3D scene ground truth was provided by the pre-designed CAD models. The camera poses ground truth can be easily accessed with perfect accuracy from the simulator.

• *Highlights*: Diverse scenes, wealthy sequences, realistic data, accurate ground truth, 3D scene ground truth.

#### *4) BONN Dynamic* [82]

Bonn Dynamic, released in 2019, is a specially designed highly dynamic RGB-D dataset collected in real world, enabling the investigation under such challenge. Both camera poses and 3D mapping ground truth are provided for evaluation.

• *Platform and sensor setup*: The dataset was captured using an RGB-D sensor (Table XII lists the properties) by a handheld carrier, and exactly in this case, motion blur could easily occur due to the 6-DoF motion pattern together with rolling shutter.

TABLE XII
THE PROPERTIES OF THE SENSORS AND GROUND TRUTH INSTRUMENTS

| Sensor/Instrument | Type | Rate | Characteristics |
|---|---|---|---|
| RGB-D Camera | ASUS Xtion Pro LIVE | 30 Hz | 1280 × 1024, RGB + Depth, rolling shutter, 58° × 45°FOV |
| Motion Capture | OptiTrack Prime 13 | 240 Hz | <0.2cm/0.5° error |
| 3D Scanner | Leica BLK360 | 360 kHz | 8mm accuracy |

• *Scenes and sequences*: A total of 24 sequences were captured under highly-dynamic conditions, such as manipulating box and playing balloon, which was known to be challenging for motion estimation, and 2 sequences were captured under static condition. Three dynamic frames are shown in Fig. 17, as can be seen, the dynamics also add effects on motion blur.

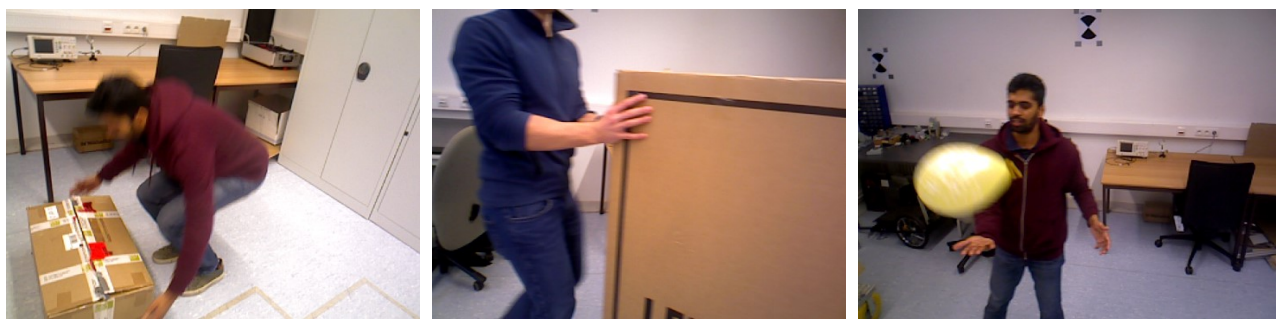

Fig. 17. Two example dynamic frames in the dataset: manipulating boxes (left), and playing with balloons (right) [82].

• *Ground truth*: The pose ground truth was measured by motion capture system with high rate and accuracy. Additionally, the 3D scene ground truth was also provided, containing only the static parts of the scene, enabling 3D reconstruction evaluation. Table XII lists the properties of the ground truth instruments.

• *Highlights*: Highly dynamic environments, wealthy sequences, accurate ground truth, 3D scene ground truth.

## D. *Image-based 3D Reconstruction Category*

### 1) *Middlebury* [13]

Middlebury, published in 2006, is a top famous multi-view stereo dataset with known 3D shape ground truth. Novel evaluation metrics were proposed to enable quantitative comparison on mapping accuracy and completeness, which has been dominantly used by the community in past decades.

• *Platform and sensor setup*: The dataset was collected using a CCD camera mounted on a Stanford Spherical Gantry robotic arm. The robotic arm has a high motion flexibility that enables the camera shooting at any viewpoints on a one-meter sphere, with a rather high positioning accuracy within 0.2mm and 0.02 degrees. The properties of the sensor are listed in Table XIII.

TABLE XIII
THE PROPERTIES OF THE SENSORS AND GROUND TRUTH INSTRUMENTS

| Sensor/Instrument | Type | Resolution | Characteristics |
|---|---|---|---|
| Camera | CCD sensor | 640 × 480 | RGB |
| 3D Scanner | Cyberware Model-15 | 0.25mm | 0.2mm accuracy |

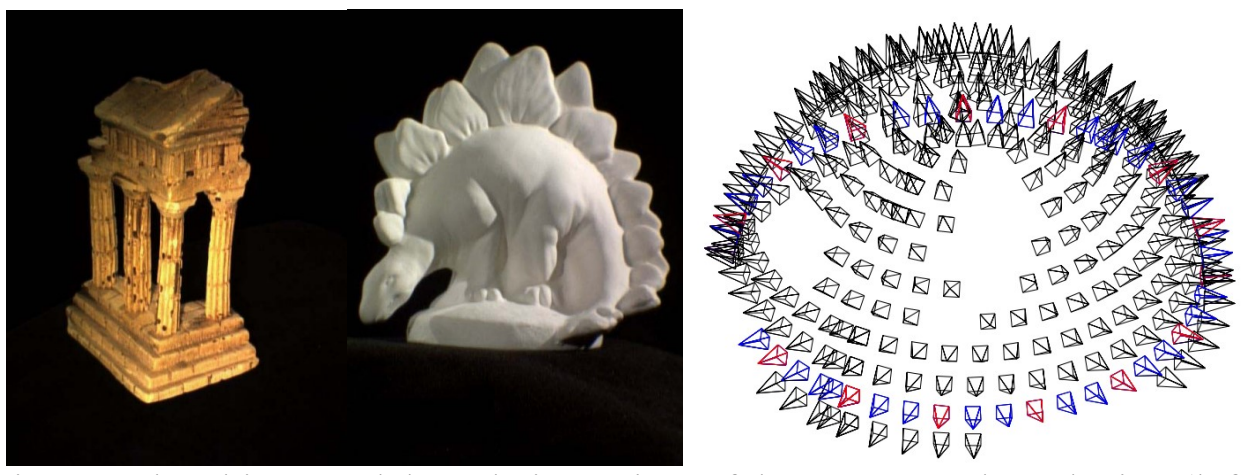

Fig. 18. The object models and viewpoints of dataset: Temple and Dino (left), and different viewpoints (right) [13].

• *Scenes and sequences*: The dataset well selected two target objects for study: Temple and Dino (as shown in the left 2 parts of Fig. 18), both of which are plaster models. Many challenging characteristics were included in the dataset, such as both sharp and smooth features, complex topologies, strong concavities, and both strong and weak surface texture. For each model, there are 3 sets of data provided (see the right part of Fig. 18): hemisphere views (black), ring views (blue), and sparse ring views (red).

• *Ground truth*: The camera poses ground truth were provided by the high-precision robotic arm. Then based on the accurate poses, for each object, about 200 individual scans acquired by a 3D scanner were merged into a whole model, which has a 0.25 mm resolution and better than 0.2mm accuracy (as listed in Table XIII), serving as 3D reconstruction ground truth.

• *Highlights*: 3D model ground truth, challenging model, novel evaluation metrics.

### 2) *BigSfM*[20]

The BigSFM Project is a gathering of massive internet photos datasets, stimulating the growth of many large-scale reconstruction methods in recent years. It offers a seminal way to get huge data resource to support large-scale 3D reconstruction study.

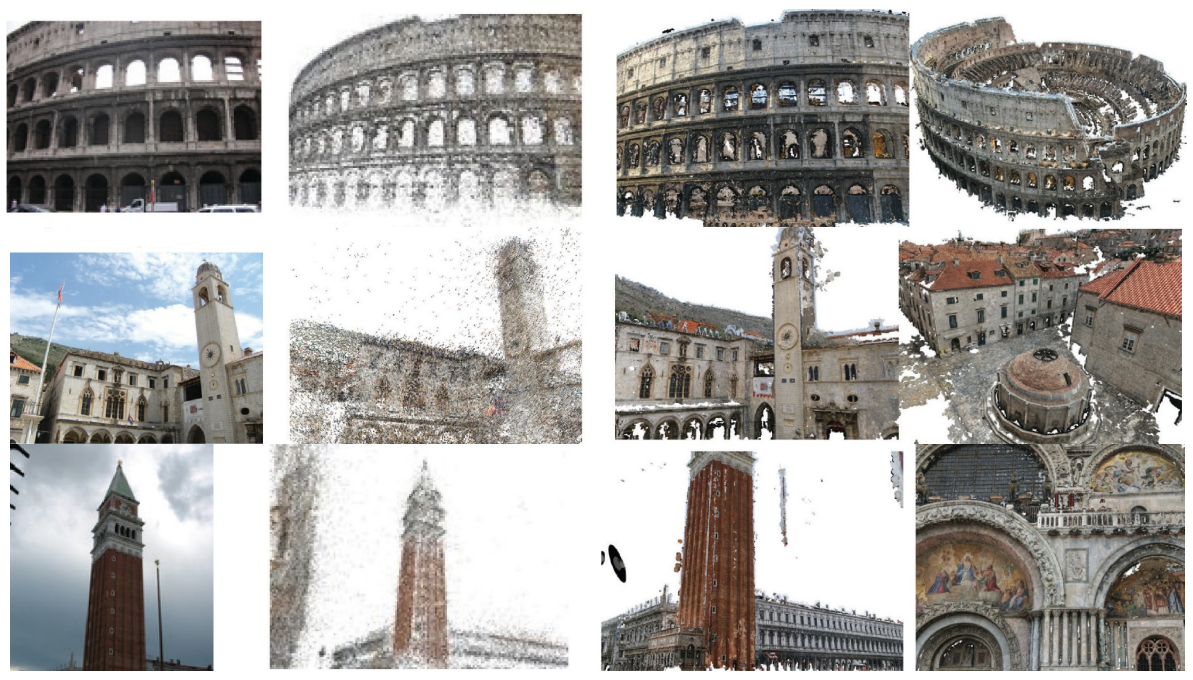

Fig. 19. Some representative images and reconstructed models of the scenes: Colosseum (Rome), Dubrovnik, and San Marco Square (Venice) from top to bound [83].

• *Platform and sensor setup*: This kind of datasets were normally collected on the internet, e.g., Flickr and Google images, thus the platforms and sensors vary. Most of the photos were captured via hand-held smartphone or digital camera, so there was not a fixed resolution or camera intrinsic, which on the other hand, can also serve as a challenge for 3D reconstruction algorithms.

• *Scenes and sequences*: BigSFM contains a variety of community photo datasets. Some famous ones are Rome [83], Venice [84], Dubrovnik [85], NotreDame [86], Landmarks [87], etc. For each scene, there could be hundreds of to tens of thousands of images shot at massive viewpoints. Some representative images and reconstructed models are shown in Fig. 19.

• *Ground truth*: Due to the unorganized nature that the photos could be collected using random cameras, by different persons, and under different conditions, it is normally impractical to provide ground truth 3D models for evaluation. As an alternative, sometimes the GPS tags along with the photos (when available) could be used to verify the camera poses, indicating the reconstruction quality, although not accurate that much.

• *Highlights*: Large-scale scenes, massive photos, web resource.

[20] https://research.cs.cornell.edu/bigsfm/

### 3) *Tanks and Temples* [88]

Tanks and Temples, published in 2017, is a large-scale image-based 3D reconstruction dataset which covers both outdoor and indoor scenes. The most significant contribution is the provided large-scale and high-quality ground truth 3D models, which were long missing in previous datasets [83], [89], [90].

• *Platform and Sensor Setup*: There are two setups for capturing the data, both using a stabilizer with a professional digital camera. As listed in Table XIV, both cameras captured 4k videos at 30hz, and the images were in high-quality thanks to the stabilizer, all of which enable high-fidelity scene reconstruction.

TABLE XIV
THE PROPERTIES OF THE SENSORS AND GROUND TRUTH INSTRUMENT

| Sensor/Instrument | Type | Rate | Characteristics |
|---|---|---|---|
| Camera #1 | DJI Zenmuse X5R | 30 Hz | 3840 × 2160, 8-bit RGB, rolling shutter, 84° FOV |
| Camera #2 | Sony a7S II | 30 Hz | 3840 × 2160, 8-bit RGB rolling shutter, 90° FOV |
| 3D Scanner | FARO Focus 3D X330 | 976 kHz | 330m range, 2mm accuracy |

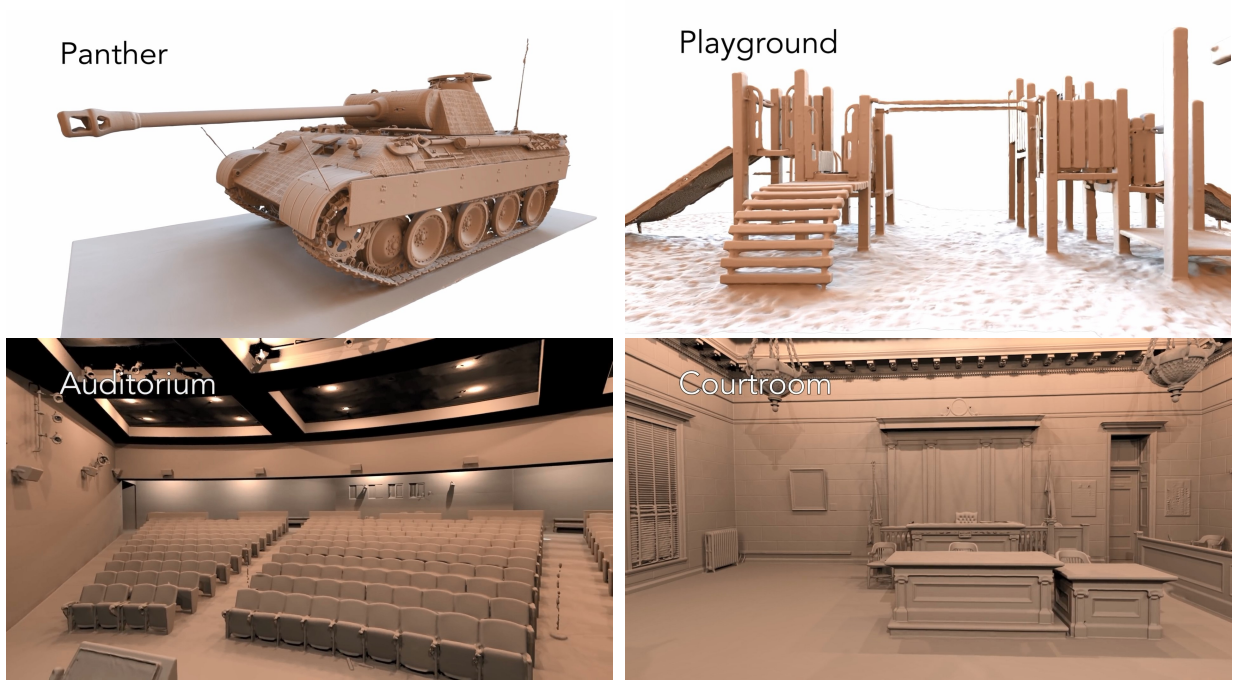

Fig. 20. 4 representative scene models: Panther and Playground - Intermediate group (up), Auditorium and Courtroom – Advanced group (bound) [91].

• *Scenes and sequences*: There is a total of 21 sequences (1 per scene) provided, categorized as 3 groups: intermediate group, advanced group, and training group, covering scenes including Panther, Playground, Auditorium, Courtroom, and so on (see Fig. 20), with an area of up to 4295m$^2$. The advanced group are more challenging due to the scale, complexity, illumination, and other factors, as can be felt from the Auditorium model.

• *Ground truth*: The 3D scene ground truth was acquired by an industrial-grade large-scale laser scanner (Table XIV lists the properties), by registering multiple individual scans together via overlap areas. Furthermore, the ground truth camera poses were estimated using the ground truth point cloud [88].

• *Highlights*: 3D scene ground truth, multiple complexity levels, large-scale scenes, diverse scenes.

## V. OVERVIEW AND COMPARISON OF EXISTING DATASETS

In this section, based on the proposed 4 broad categories, we comprehensively survey 97 SLAM related datasets, and provide concise but comprehensive overview and comparison tables by category (in a chronological order, for several datasets proposed in the same year, we follow the alphabet order) to serve as a dictionary. We cover the 5 fundamental dimensions of datasets introduced in Section II, and clearly specify the applicable tasks and challenging elements of each dataset. For each category of datasets, two tables are provided: in the first table, except for the dataset name and released/published year, we arrange cited number, perception sensor setup, ground truth, applicable tasks, and tested methods, for a general selection; in the second table, we arrange mobile platform, collection scenes, data sequences, challenging elements, and dataset links, enabling a finer selection and easy access. The details of each dimension are illustrated below.

• *Name*: The dataset names (usually from the providers) and the corresponding references. If a dataset is collected by synthetic and virtual manner, then we will annotate it as “V”. If a dataset is collected from the web, then we will annotate it as “W”.

• *Year*: The released or published year of the datasets.

• *Cited Number*: The citation number of the dataset (we refer to the statistics on Google Scholar), which could partly indicate the utilization. We use “partly” here because: on the one hand, many datasets do not have self-contained publications [83], then the citation shares together with the corresponding algorithm papers; on the other hand, some datasets are not specifically designed for SLAM related tasks [92], thus the citation could be contributed by other usages.

• *Perception Sensor Setup*: This dimension lists the calibrated perception sensors, framing into 3 columns: camera (we also put RGB-D sensor into this class), IMU, and LiDAR. The camera column lists the specific camera types, such as color, grayscale, event, depth, RGB-D, plenoptic, thermal, and intensity, along with some features to be emphasized such as stereo, fisheye, omnidirectional, wide-angle (WA), narrow-angle (NA), global shutter (GS), and rolling shutter (RS). For image-based 3D reconstruction datasets, we point out either the camera type is professional, consumer, or random to indicate the image source and quality. The IMU column shows whether perception purpose IMU is available, that means we will not cover an IMU that mainly used for ground truth generation. The LiDAR column shows the LiDAR type (2D or 3D) and number of beam (e.g., 4, 16, 32, and 64). For the very few datasets with Radar, we also put it in the LiDAR column, since they are all ranging sensors.

• *Ground truth*: This dimension is framed into 2 columns: pose and 3D. Each column shows whether such type of ground truth is available, and if available the generation technique will be shown, indicating the accuracy.

• *Applicable tasks*: This dimension clearly specifies the applicable tasks of each dataset. For vision-based, LiDAR-related, and RGB-D-based categories, the tasks are framed into 4 columns: O, OAM, SLAM, and 3DR. If the pose ground truth is available, then we will give a “✓” in the blank of O, OAM, and SLAM; if the 3D ground truth is available, then we will give a “✓” in 3DR. In addition, based on the definition of the proposed taxonomy, SLAM has loop closure or relocalization features, hence to well support them, we will give another “✓” if the sequences contain revisited places. However, we will not give a “✓” if the loop is used for evaluation, because SLAM can close the loop and bias the result. For image-based category, 2 columns are listed: SfM and MVS. We will give a “✓” to SfM and MVS first, as it is already a success to reconstruct to 3D from massive individual 2D images. And if either pose or 3D ground truth is available, we will give another "✓" in the blank of SfM; if the 3D ground truth is available, we will give another "✓" to MVS.

- *Tested methods*: This dimension lists some tested methods on corresponding datasets, in the hope of indicating the usage status and providing useful references. Limited by the space, we cannot list all these contents, which can be further indexed via database. IMU fusion methods (e.g., VINS-mono [102]) are underlined.
- *Mobile Platform*: In this dimension, the mobile platform of collection is given, indicating different motion patterns.
- *Collection Scene*: This dimension shows the collection scenes of data sequences, framed into 3 columns: indoor, outdoor, and indoor-outdoor integration (I-O) scenes. The indoor and outdoor columns list the specific scenes, such as urban, office, room, and so on. If there are many types of scenes employed, we will fill the blank as various or multiple. The I-O column reveals whether the data sequence is traversed between indoor and outdoor scenes (if yes, we will give a "✓" in the blank).
- *Data Sequence*: This dimension is framed into 2 columns: number and volume. The number column lists the total sequence number of the datasets, indicating the variety of collection routes. To avoid misunderstanding, the dataset will be annotated a "r" if the sequences have repeated routes. The volume column shows the length, duration, or frame number of the data sequences.
- *Challenging element*: In this dimension, challenging elements within the datasets are listed, which could be resulted by motion pattern, scene type, and scene conditions. Overall, elements caused by motion pattern will be annotated with an "M", mainly including rapid motion, pure rotation, motion blur, and LiDAR distortion; for elements resulted by scene type, namely weak, repetitive, and interference of feature and structure, we will mark as "F" and "S" respectively; for scene conditions, we will mark dynamic objects as "D", varying, bad, and changing illumination as "I", various and adverse weathers as "W", different time-slots as "T", multiple seasons as "SE". For image-based datasets, we will additionally mark complex geometry as "C", and large-scale sequences (more than 1k frames in each) as "L".
- *Dataset Link*: The web links of datasets. One can easily access detailed information and download the datasets via the links.

**** Dataset Selection Guideline***

We recommend the dataset users to follow our guideline to select useful datasets using our "dictionary" (the following tables).

1) Determine which broad category your algorithm belongs to (refer to Section III) and go to the corresponding session (A~D). E.g., if you are building vision-based SLAM related algorithms, then go to session A, B, and C.
2) Consider the specific task you are building (refer to Section III), and see the applicable tasks columns (O/OAM/SLAM/3DR or SfM/MVS) to find available datasets (normally you can find many candidates in this step).
3) Consider the specific sensors you want to use (camera, stereo, event, IMU, LiDAR, etc.), and see the perception sensor setup columns to furtherly find some suitable datasets. E.g., if you want to build stereo-vision algorithm with IMU fusion, you can locate EuRoC, Blackbird, Complex Urban, etc.; if you want event-based vision datasets, you can locate UZH-event, MVSEC, etc. Note that, some datasets with IMU data and 6-DoF motions could be with higher difficulty.
4) Consider the indicator you want to evaluate (pose or 3D structure), and see the ground truth columns to choose datasets with expected ground truth quality (refer to Section II-B-5)). E.g., if you want to study the frame-by-frame accuracy of VO, then better to choose those datasets with MoCap or LasTrack.
5) Choose your expected mobile platform (they are with different motion pattern, refer to Section II).
6) Choose your expected testing scene (indoor, outdoor, indoor-outdoor, or specifically office, campus...).
7) Choose your desired data volume (100 frames or thousands of frames per sequence).
8) Choose your expected challenging elements within the datasets (weak texture, bad illumination, adverse weather, and so on).
9) Go to the dataset link to get detailed information and download for use.

*Notice*: Except for "demand-oriented" selection, from the algorithm testing point of view, one had better choose multiple datasets (with different platform, scene, etc.) for evaluation to avoid over-fitting and hidden defects on certain benchmarks. We will later analyze the necessity of complementary usage in Section VI-C.

## *A. Vision-based Mobile Localization and Mapping and 3D Reconstruction Category*

In this part, we collect 34 datasets for vision-based mobile localization and mapping and 3D reconstruction tasks. The overview and comparison are given in Table XV and Table XVI. Note that we also include several datasets that equipped with LiDAR sensors, such as OpenLORIS and TartanAir, because such datasets are mainly designed for vision-based algorithms, and the LiDARs are mainly employed to generate ground truth [41] or merely serve as a kind of ground truth label [72].

TABLE XV
OVERVIEW AND COMPARISON TABLE-I OF VISION-BASED MOBILE LOCALIZATION AND MAPPING AND 3D RECONSTRUCTION DATASETS

| Name | Year | Cited | Perception Sensor Setup | | | Ground Truth | | Applicable Tasks | | | | Tested Methods |
|---|---|---|---|---|---|---|---|---|---|---|---|---|
| | | | Camera | IMU | LiDAR | Pose | 3D | O | OAM | SLAM | 3DR | |
| UTIAS MultiRobot [93] | 2011 | 56 | 1×color (per-robot) | N/A | N/A | MoCap[a] | N/A | ✓ | ✓ | ✓✓ | | [94], [95] |
| Devon Island [96] | 2012 | 55 | 2×color (stereo) | N/A | N/A | DGPS | N/A | ✓ | ✓ | ✓ | | [97], [98] |
| TUM Omni [99] | 2015 | 206 | 1×fisheye-gray | N/A | N/A | MoCap (P[b]) | N/A | ✓ | ✓ | ✓✓ | | [99], [100] |
| EuRoC MAV [15] | 2016 | 682 | 2×gray (stereo) | Y | N/A | MoCap<br>LasTrack[c] | 3D Scanner | ✓ | ✓ | ✓✓ | ✓ | [49], [53], [101], [102] |
| (V) SYNTHIA [92] | 2016 | 1031 | 2×omni-color (stereo)<br>2×omni-depth (stereo) | N/A | N/A | Simulated | N/A | ✓ | ✓ | ✓✓ | | [103], [104] |
| TUM monoVO [38] | 2016 | 135 | 1×NA-gray<br>1×WA-gray | N/A | N/A | Loop Drift | N/A | ✓ | ✓ | ✓ | | [98], [101] |
| (V) Virtual KITTI [105] | 2016 | 628 | 1×color<br>1×depth | N/A | N/A | Simulated | N/A | ✓ | ✓ | ✓✓ | | [103], [106] |
| PennCOSYVIO [107] | 2017 | 57 | 4×color (4-orien[c])<br>2×gray (stereo)<br>1×fisheye-gray | Y | N/A | Fuducial Marker | N/A | ✓ | ✓ | ✓✓ | | [102], [108] |
| UZH Event [109] | 2017 | 204 | 1×event | Y | N/A | MoCap (P) | N/A | ✓ | ✓ | ✓✓ | | [50], [110], [124] |
| Zurich Urban [111] | 2017 | 63 | 1×color | Y | N/A | GPS+Pix4D | N/A | ✓ | ✓ | ✓✓ | | |
| ADVIO [112] | 2018 | 31 | 1×color<br>1×gray | Y | N/A | Tango+IMU+<br>Fixed-points | N/A | ✓ | ✓ | ✓✓ | | [49], [101], [113], [114] |
| Apollo Scapes [115] | 2018 | 36 | 2×color | N/A | 2×Rigel-2D | RTK-GNSS | MMS | ✓ | ✓ | ✓✓ | ✓ | [115], [116] |
| (V) Blackbird [117] | 2018 | 24 | 2×gray (stereo)<br>1×gray | Y | N/A | MoCap | N/A | ✓ | ✓ | ✓✓ | | [49], [102], [118] |
| VI Canoe [119] | 2018 | 14 | 2×color (stereo) | Y | N/A | GPS&INS | N/A | ✓ | ✓ | ✓✓ | | [120] |
| Mount Etna [121] | 2018 | 11 | 2×gray (stereo) | Y | N/A | DGPS | N/A | ✓ | ✓ | ✓✓ | | [122] |
| MVSEC [123] | 2018 | 91 | 2×event (stereo)<br>2×gray (stereo) | Y | 1×Velodyne-16 | MoCap/<br>LiDAR+IMU+GPS | N/A | ✓ | ✓ | ✓✓ | | [124] |
| SPO [125] | 2018 | 5 | 2×gray (stereo)<br>1×plenoptic | N/A | N/A | Loop Drift | N/A | ✓ | ✓ | ✓ | | [49], [101] |
| TUM VI [126] | 2018 | 90 | 2×gray (stereo) | Y | N/A | MoCap (P) | N/A | ✓ | ✓ | ✓✓ | | [53], [102], [113] |
| Upenn Fast Flight [127] | 2018 | 190 | 2×gray (stereo) | Y | N/A | GPS | N/A | ✓ | ✓ | ✓✓ | | [102], [128] |
| AQUALOC [129] | 2019 | 13 | 2×gray<br>(1 for each scene) | Y | N/A | COLMAP | N/A | ✓ | ✓ | ✓✓ | | [53], [130] |
| Dynamic Scenes [131] | 2019 | 5 | 2×color (stereo) | N/A | N/A | Loop Drifts | N/A | ✓ | ✓ | ✓✓ | | |
| OpenLORIS [41] | 2019 | 21 | 1×RGB-D<br>2×fisheye-RGB (stereo) | Y | 1×Hokuyo-2D<br>1×Robosense-16 | MoCap/<br>LiDAR SLAM | N/A | ✓ | ✓ | ✓✓ | | [49], [101], [102] |
| Rosario [132] | 2019 | 14 | 2×color (stereo) | Y | N/A | RTK-GPS | N/A | ✓ | ✓ | ✓✓ | | [49], [127], [133] |
| TUM RS [134] | 2019 | 6 | 1×GS-gray<br>1×RS-gray | Y | N/A | MoCap | N/A | ✓ | ✓ | ✓✓ | | [134], [135] |
| UZH-FPV [70] | 2019 | 47 | 1×event<br>2×gray (stereo) | Y | N/A | LasTrack | N/A | ✓ | ✓ | ✓✓ | | [50], [102] |
| ViViD [136] | 2019 | 1 | 1×thermal<br>1×RGB-D<br>1×event | Y | 1×Velodyne-16 | MoCap/<br>LiDAR SLAM | N/A | ✓ | ✓ | ✓✓ | | |
| ZJU&SenseTime [137] | 2019 | 12 | 1×color (per scene) | Y | N/A | MoCap | N/A | ✓ | ✓ | ✓✓ | | [49], [101], [102], [128] |
| 4-Seasons [138] | 2020 | 6 | 2×gray (stereo) | Y | N/A | RTK-GNSS+SLAM | SLAM | ✓ | ✓ | ✓✓ | ✓ | |
| FinnForest [140] | 2020 | 2 | 4×color<br>(2 for stereo) | Y | N/A | GNSS&INS | N/A | ✓ | ✓ | ✓✓ | | [49], [133] |
| PanoraMIS [141] | 2020 | 0 | 1×omni-color/<br>1×dual-fisheye-color | Y | N/A | Robot Odometry/<br>GPS | SLAM | ✓ | ✓ | ✓✓ | ✓ | |
| (V) TartanAir [72] | 2020 | 11 | 2×color (stereo)<br>1×depth | Y | 1×Simulated-32 | Simulated | Simulated | ✓ | ✓ | ✓✓ | ✓ | [49], [101] |
| (V) Virtual KITTI2 [142] | 2020 | 17 | 2×color (stereo)<br>1×depth | N/A | N/A | Simulated | N/A | ✓ | ✓ | ✓✓ | | |
| DSEC [143] | 2021 | 8 | 2×color (stereo)<br>2×event (stereo) | N/A | 1×Velodyne-16 | RTK-GNSS | N/A | ✓ | ✓ | ✓✓ | | [124] |
| SeRM [144] | 2021 | 0 | 1×color | N/A | N/A | RTK-GPS+<br>Wheel encoder | N/A | ✓ | ✓ | ✓✓ | | |

[a]MoCap means Motion Capture System, [b]P means partially coverage, [c]LasTrack means Laser Tracker. Within the tested methods, [49] denotes ORB-SLAM2, [50] Ultimate SLAM, [53] ORB-SLAM3 (dot-lined if both w/ and w/o IMU are tested), [101] DSO, [102] VINS-mono, [113] ROVIO, [124] ESVO, [128] MSCKF.

TABLE XVI
OVERVIEW AND COMPARISON TABLE-II OF VISION-BASED MOBILE LOCALIZATION AND MAPPING AND 3D RECONSTRUCTION DATASETS

| Name | Mobile Platform | Collection Scene | | | Data Sequence | | Challenging | Dataset Link |
|---|---|---|---|---|---|---|---|---|
| | | Indoor | Outdoor | I-O | Number | Volume | | |
| UTIAS MultiRobot | Wheeled Robot | Workspace | | | 9 | 900-4400s per seq | | http://asrl.utias.utoronto.ca/datasets/mrclam/index.html |
| Devon Island | Rover | | Island (planetary) | | 23 | 10km in total | F, T | http://asrl.utias.utoronto.ca/datasets/devon-island-rover-navigation/ |
| TUM Omni | Handheld | Office | | | 5 | 79-165s per seq | M | https://vision.in.tum.de/data/datasets/omni-lsdslam |
| EuRoC MAV | UAV | Machine hall, room | | | 11 | 37-131m per seq | M, I | https://projects.asl.ethz.ch/datasets/doku.php?id=kmavvisualinertialdatasets |
| (V) SYNTHIA | Car | | Urban | | 51(r) | 1k frames per seq | D, T, W, SE | https://synthia-dataset.net/ |
| TUM monoVO | Handheld | Building | Campus | ✓ | 50 | 105min in total | M | https://vision.in.tum.de/data/datasets/mono-dataset |
| (V) Virtual KITTI | Car | | Urban | | 50(r) | 233-837 frames per seq | W, T | https://europe.naverlabs.com/research/computer-vision-research-naver-labs-europe/proxy-virtual-worlds-vkitti-1/ |
| PennCOSYVIO | Handheld | Building | Campus | ✓ | 4 | 150m per seq | F, I | https://daniilidis-group.github.io/penncosyvio/ |
| UZH Event | Handheld | Objects, office | Building | | 24 | 2-133s per seq | | http://rpg.ifi.uzh.ch/davis_data.html |
| Zurich Urban | UAV | | Urban | | 1 | 2km in total | | http://rpg.ifi.uzh.ch/zurichmavdataset.html |
| ADVIO | Handheld | Office, mall, metro | Urban, campus | | 7 | 4.5km in total | M, F | https://github.com/AaltoVision/ADVIO |
| ApolloScape | Car | | Urban | | 6 | 28km in total | D, T, I | http://apolloscape.auto/self_localization.html |
| (V) Blackbird | UAV | Various | | | 168(r) | 860.8m max | M | https://github.com/mit-fast/Blackbird-Dataset |
| VI Canoe | Canoe | | River | | 10 | 2.7km in total | F | https://databank.illinois.edu/datasets/IDB-9342111 |
| Mount Etna | Rover | | Mountain | | 2 | 1km per seq | F | https://www.robex-allianz.de/datasets/ |
| MVSEC | Multiple | Workspace, building | Workspace, building, urban, highway | ✓ | 14 | 9.8-18293m per seq | I | https://daniilidis-group.github.io/mvsec/ |
| SPO | Handheld | Building | Park | | 11 | 25-274m per seq | | https://www.hs-karlsruhe.de/odometry-data/ |
| TUM VI | Handheld | Building | Campus | ✓ | 28 | 67-2656m per seq | F, I | https://vision.in.tum.de/data/datasets/visual-inertial-dataset |
| Upenn Fast Flight | UAV | Warehouse | Wooded area | ✓ | 4(r) | 700m per seq | M, F, I | https://github.com/KumarRobotics/msckf_vio |
| AQUALOC | Underwater vehicle | | Underwater | | 17 | 10.7-122.1m per seq | F, I, D | http://www.lirmm.fr/aqualoc/ |
| Dynamic Scenes | Car | | Urban | | 2 | 6.4k frames per seq | D, I | https://dataverse.harvard.edu/dataset.xhtml?persistentId=doi:10.7910/DVN/NZETVT |
| OpenLORIS | Wheeled Robot | Multiple | | | 22(r) | 2244s in total | I, T, D | https://lifelong-robotic-vision.github.io/dataset/scene |
| Rosario | Wheeled Robot | | Agriculture | | 6 | 152-709m per seq | F, I | https://www.cifasis-conicet.gov.ar/robot/doku.php |
| TUM RS | Handheld | Office | | | 10 | 30-57m per seq | M | https://vision.in.tum.de/data/datasets/rolling-shutter-dataset |
| UZH-FPV | UAV | Hangar | Field | | 27 | 59-736m per seq | M | http://rpg.ifi.uzh.ch/uzh-fpv.html |
| ViViD | Handheld | Room | Urban | | 9 | Up to 50m per seq | M, I, T | https://sites.google.com/view/dgbicra2019-vivid |
| ZJU&SenseTime | Handheld | Room | | | 16 | 40-119s per seq | M, F, D | http://www.zjucvg.net/eval-vislam/ |
| 4-Seasons | Car | | Urban, rural, sub-urban | | 30 | 350km in total | I, T, W, D, SE | https://www.4seasons-dataset.com/ |
| FinnForest | Car | | Forest | | 11 | 1.3-6.5km per seq | F, I, W, T, SE | https://etsin.fairdata.fi/dataset/06926f4b-b36a-4d6e-873c-aa3e7d84ab49 |
| PanoraMIS | Wheeled Robot/ UAV | | Urban, park, field | | 4 | Up to 4.2km per seq | F, I | https://home.mis.u-picardie.fr/~panoramis/ |
| (V) TartanAir | Random | Various | Various | | 1037 | 1M frames in total | I, D, W, SE | https://theairlab.org/tartanair-dataset/ |
| (V) Virtual KITTI2 | Car | | Urban | | 50(r) | 233-837 frames per seq | W, T | https://europe.naverlabs.com/research/computer-vision-research-naver-labs-europe/proxy-virtual-worlds-vkitti-2/ |
| DSEC | Car | | Urban | | 53 | 3193s in total | I, T | https://dsec.ifi.uzh.ch/ |
| SeRM | Car | | Urban | | 3 | 26km in total | D | https://github.com/SeRM-dataset/dataset |

## *B. LiDAR-related Mobile Localization And Mapping And 3D Reconstruction Category*

In this part, we collect 25 datasets for LiDAR-related mobile localization and mapping and 3D reconstruction usages. The overview and comparison are given in Table XVII and Table XVIII.

TABLE XVII
OVERVIEW AND COMPARISON TABLE-I OF LIDAR-RELATED MOBILE LOCALIZATION AND MAPPING AND 3D RECONSTRUCTION DATASETS

| Name | Year | Cited | Perception Sensor Setup | | | Ground Truth | | Applicable Tasks | | | | Tested Methods |
|---|---|---|---|---|---|---|---|---|---|---|---|---|
| | | | Camera | IMU | LiDAR | Pose | 3D | O | OAM | SLAM | 3DR | |
| DARPA [145] | 2007 | 58 | 4×color (stereo)<br>1×WA-color | N/A | 12×Sick-2D<br>1×Velodyne-64 | DGPS+IMU+DMI | N/A | ✓ | ✓ | ✓✓ | | |
| New College [78] | 2009 | 308 | 2×gray (stereo)<br>5×color (panor[a]) | Y | 2×Sick-2D | Segway Odometry | N/A | ✓ | ✓ | ✓✓ | | [146], [147] |
| Rawseeds [148] | 2009 | 153 | 3×gray (stereo)<br>1×color<br>1×omni[b]-color | Y | 2×Hokuyo-2D<br>2×Sick-2D | VisualTag/<br>LiDAR-ICP/<br>RTK-GPS | N/A | ✓ | ✓ | ✓✓ | | [149] |
| Marulan [150] | 2010 | 78 | 1×color<br>1×IR-thermal | N/A | 4×Sick-2D | RTK-DGPS&INS | N/A | ✓ | ✓ | ✓✓ | | [151] |
| Ford Campus [152] | 2011 | 253 | 6×color (omni) | Y | 1×Velodyne-64<br>2×Riegl-2D | DGPS | N/A | ✓ | ✓ | ✓✓ | | [153], [154] |
| ASL Challenging [155] | 2012 | 102 | N/A | Y | 1×Hokuyo-2D<br>(tilting as 3D) | Total Station | N/A | ✓ | ✓ | ✓✓ | | [153], [156] |
| Gravel Pit [157] | 2012 | 12 | 1×intensity (LiDAR) | N/A | 1×Autonosys-3D | DGPS | N/A | ✓ | ✓ | ✓✓ | | [158] |
| KITTI [14] | 2012 | 6341 | 2×color (stereo)<br>2×gray (stereo) | N/A | 1×Velodyne-64 | RTK-GPS&INS | N/A | ✓ | ✓ | ✓✓ | | [45], [49], [51], [53], [54], [154] |
| Canadian Planetary [159] | 2013 | 40 | 3×color (stereo) | Y (P[c]) | 1×Sick-2D | DGPS+IMU/<br>DGPS+VO | N/A | ✓ | ✓ | ✓✓ | | [160] |
| Institut Pascal [161] | 2013 | 20 | 4×gray (2 for stereo)<br>1×omni-color<br>1×fisheye-color<br>1×webcam-color | Y | 2×Sick-2D | RTK-GPS | N/A | ✓ | ✓ | ✓✓ | | [162] |
| Malaga Urban [163] | 2014 | 178 | 2×color (stereo) | Y | 2×Sick-2D<br>3×Hokuyo-2D | GPS | N/A | ✓ | ✓ | ✓✓ | | [49], [164] |
| NCLT [165] | 2016 | 166 | 6×color (omni) | Y | 1×Velodyne-32<br>1×Hokuyo-2D | FOG+RTK-GPS+<br>EKF-Odometry | N/A | ✓ | ✓ | ✓✓ | | [166], [167] |
| Sugar Beets 2016 [168] | 2016 | 75 | 1×RGB/NIR<br>1×RGB-D | N/A | 2×Velodyne-16<br>1×Nippon-2D | RTK-GPS | 3D Scanner | ✓ | ✓ | ✓✓ | ✓ | [169] |
| Chilean Underground [170] | 2017 | 16 | 3×color (stereo) | N/A | 1×Rigel-3D | Registration | 3D Scanner | ✓ | ✓ | ✓✓ | ✓ | |
| Oxford RobotCar [40] | 2017 | 631 | 3×color (stereo)<br>3×fisheye-color | N/A | 2×Sick-2D<br>1×Sick-4 | RTK-GPS&INS | N/A | ✓ | ✓ | ✓✓ | | [48], [49], [171], [172] |
| KAIST Day/Night [173] | 2018 | 63 | 2×color (stereo)<br>1×thermal | N/A | 1×Velodyne-32 | RTK-GPS&INS | N/A | ✓ | ✓ | ✓✓ | | [174] |
| Katwijk Beach [175] | 2018 | 17 | 2×color (stereo)<br>2×color (stereo) | N/A | 1×Velodyne-16 | RTK-DGPS | N/A | ✓ | ✓ | ✓✓ | | |
| MI3DMAP [176][177] | 2018 | 29 | 1×color<br>2×fisheye-color | Y | 2×Velodyne-16 | SLAM | 3D Scanner | ✓ | ✓ | ✓✓ | ✓ | [178] |
| Urban@CRAS [179] | 2018 | 12 | 2×color (stereo) | Y | 1×Velodyne-16 | RTK-GPS | N/A | ✓ | ✓ | ✓✓ | | [49], [133] |
| Complex Urban [37] | 2019 | 50 | 2×color (stereo) | Y | 2×Velodyne-16<br>2×Sick-2D | RTK-GPS+<br>FOG+SLAM | SLAM | ✓ | ✓ | ✓✓ | ✓ | [172] |
| Mulran [180] | 2019 | 17 | N/A | N/A | 1×Radar<br>1×Sick-2D<br>1×Velodyne-16 | RTK-GPS+<br>FOG+SLAM | SLAM | ✓ | ✓ | ✓✓ | ✓ | [181] |
| Newer College [77] | 2020 | 6 | 1×RGB-D<br>(used as binocular) | Y | 1×Ouster-64 | ICP Map Matching | 3D Scanner | ✓ | ✓ | ✓✓ | ✓ | [49], [182] |
| Oxford Radar [184] | 2020 | 58 | 3×color (stereo)<br>3×fisheye-color | N/A | 1×Radar<br>2×Sick-2D<br>2×Velodyne-32 | GPS&INS+SLAM | N/A | ✓ | ✓ | ✓✓ | | [185], [192] |
| UrbanLoco [186] | 2020 | 14 | 1×fisheye-color/<br>6×RGB | Y | 1×Velodyne-32/<br>1×Robosense-32 | RTK-GNSS&INS | N/A | ✓ | ✓ | ✓✓ | | [45], [102] |
| Naver Labs Indoor* [187] | 2021 | 1 | 6×color<br>4×color (smartphone) | N/A | 2×Velodyne-16 | SLAM+SfM | SLAM+SfM | ✓ | ✓ | ✓✓ | ✓ | |

[a]panor means panoramic, [b]omni means omnidirectional, [c]P means partially coverage. *Naver Labs has released 2 datasets, whereas the outdoor dataset is only opened to Korean users, thus has not been listed here. Within the tested methods, [45] denotes LOAM, [48] LSD-SLAM, [49] ORB-SLAM2, [51] LeGO-LOAM, [53] ORB-SLAM3, [54] V-LOAM, [102] VINS-mono, [146] ORB-SLAM, [154] OverlapNet.

TABLE XVIII
OVERVIEW AND COMPARISON TABLE-II OF LIDAR-RELATED MOBILE LOCALIZATION AND MAPPING AND 3D RECONSTRUCTION DATASETS

| Name | Mobile Platform | Collection Scene | | | Data Sequence | | Challenging | Dataset Link |
|---|---|---|---|---|---|---|---|---|
| | | Indoor | Outdoor | I-O | Number | Volume | | |
| DARPA | Car | | Urban | ✓ | 3 | 90km in total | D, I | http://grandchallenge.mit.edu/wiki/index.php/PublicData |
| New College | Wheeled Robot | | Campus | | 1 | 2.2km in total | I | http://www.robots.ox.ac.uk/NewCollegeData |
| Rawseeds | Wheeled Robot | Building | Campus | ✓ | 11 | 10km in total | D | http://www.rawseeds.org/ |
| Marulan | UGV | | Field | | 40 | 66s-357s per seq | W, T | http://sdi.acfr.usyd.edu.au/ |
| Ford Campus | Car | | Campus | | 2 | 6km and 37km | | http://robots.engin.umich.edu/SoftwareData/Ford |
| ASL Challenging | Custom Base | Multiple | Multiple | | 8 | 31-45 scans per seq | S | https://projects.asl.ethz.ch/datasets/doku.php?id=laserregistration:laserregistration |
| Gravel Pit | Rover | | Field (planetary) | | 10(r) | 1.1km per seq | F, S, T | http://asrl.utias.utoronto.ca/datasets/abl-sudbury/ |
| KITTI | Car | | Urban, sub-urban | | 22 | 39.2km in total | D | http://www.cvlibs.net/datasets/kitti/index.php |
| Canadian Planetary | UGV | Workspace (planetary) | Workspace (planetary) | | 4 | 272 scans in total | F, S | http://asrl.utias.utoronto.ca/datasets/3dmap/#Datasets |
| Institut Pascal | Car | | Workspace | | 6 | 0.6-7.8km per seq | | http://ipds.univ-bpclermont.fr/ |
| Malaga Urban | Car | | Urban | | 1 | 36.8km in total | D | https://www.mrpt.org/MalagaUrbanDataset |
| NCLT | Wheeled Robot | Building | Campus | ✓ | 27 | 147.4km in total | D, T, I, W, SE | http://robots.engin.umich.edu/nclt/ |
| Sugar Beets 2016 | Field Robot | | Agriculture | | 27 | 2-5km per seq | F, S, T | http://www.ipb.uni-bonn.de/data/sugarbeets2016/ |
| Chilean Underground | UGV | Mine | | | 43 | 2km in total | I, S | http://dataset.amtc.cl/ |
| Oxford RobotCar | Car | | Urban | | 100(r) | 10km per seq | D, I, T, W, SE | https://robotcar-dataset.robots.ox.ac.uk/ |
| KAIST Day/Night | Car | | Campus, urban, residential | | 18(r) | 42km in total | I, T, D | https://sites.google.com/view/multispectral/home |
| Katwijk Beach | Rover | | Beach (planetary) | | 3 | 2km in total | I | https://robotics.estec.esa.int/datasets/katwijk-beach-11-2015/ |
| MI3DMAP | Backpack | Building | | | 8 | Up to 500m per seq | | http://mi3dmap.net/ |
| Urban@CRAS | Car | | Urban | | 5 | 1.8-5km per seq | I, D | https://rdm.inesctec.pt/dataset/nis-2018-001 |
| Complex Urban | Car | | Urban, campus | | 19 | 1.6-22km per seq | D | https://sites.google.com/view/complex-urban-dataset |
| Mulran | Car | | Urban | | 12 | 4.9-23.4km per seq | D, T | https://sites.google.com/view/mulran-pr/home |
| Newer College | Handheld | | Campus | | 1 | 2.2km in total | | https://ori-drs.github.io/newer-college-dataset/ |
| Oxford Radar | Car | | Urban | | 32 | 9km per seq | I, D, T, W | https://oxford-robotics-institute.github.io/radar-robotcar-dataset/ |
| UrbanLoco | Car | | Urban | | 13 | Up to 13.8km per seq | D | https://advdataset2019.wixsite.com/urbanloco |
| Naver Labs Indoor | Wheeled Robot | Mall, metro | | | 5 | ~137k frames in total | F, D, I, T | https://www.naverlabs.com/en/storyDetail/194 |

## C. RGB-D-based Mobile Localization And Mapping And 3D Reconstruction Category

In this part, we collect 20 datasets for RGB-D-based mobile localization and mapping and 3D reconstruction tasks. The overview and comparison are given in Table XIX and Table XX.

TABLE XIX
OVERVIEW AND COMPARISON TABLE-I OF RGB-D-BASED MOBILE LOCALIZATION AND MAPPING AND 3D RECONSTRUCTION DATASETS

| Name | Year | Cited | Perception Sensor Setup | | | Ground Truth | | Applicable Tasks | | | | Tested Methods |
|---|---|---|---|---|---|---|---|---|---|---|---|---|
| | | | Camera | IMU | LiDAR | Pose | 3D | O | OAM | SLAM | 3DR | |
| KinectFusion [189] | 2012 | 82 | 1×RGB-D | N/A | N/A | SLAM | SLAM/3D Scanner | ✓ | ✓ | ✓✓ | ✓ | [55] |
| TUM RGB-D [12] | 2012 | 2131 | 1×RGB-D | N/A | N/A | MoCap[a] | N/A | ✓ | ✓ | ✓✓ | | [48], [49], [146], [190], [212] |
| 7-Scenes [191] | 2013 | 477 | 1×RGB-D | N/A | N/A | SLAM | SLAM | ✓ | ✓ | ✓✓ | ✓ | [192] |
| SUN 3D [193] | 2013 | 511 | 1×RGB-D | N/A | N/A | SfM | SfM | ✓ | ✓ | ✓✓ | ✓ | [80], [193], [194], [240] |
| Stanford 3D Scene [195] | 2013 | 181 | 1×RGB-D | N/A | N/A | SLAM | SLAM | ✓ | ✓ | ✓✓ | ✓ | [195] |
| (V) ICL-NUIM [79] | 2014 | 626 | 1×RGB-D | N/A | N/A | SLAM | Simulated | ✓ | ✓ | ✓✓ | ✓ | [56], [80], [190], [194], [240] |
| RGB-D Scenes [196] | 2014 | 294 | 1×RGB-D | N/A | N/A | SLAM | SLAM | ✓ | ✓ | ✓✓ | ✓ | |
| CoRBS [197] | 2016 | 79 | 1×RGB-D | N/A | N/A | MoCap | 3D Scanner | ✓ | ✓ | ✓✓ | ✓ | [198], [240] |
| SceneNN [199] | 2016 | 152 | 1×RGB-D (per scene) | N/A | N/A | SLAM | SLAM | ✓ | ✓ | ✓✓ | ✓ | [200] |
| ETH RGB-D [201] | 2017 | 153 | 1×RGB-D | N/A | N/A | MoCap | 3D Scanner | ✓ | ✓ | ✓✓ | ✓ | [201] |
| Robot@Home [202] | 2017 | 49 | 4×RGB-D | N/A | 1×Hokuyo-2D | SLAM | N/A | ✓ | ✓ | ✓✓ | | [203] |
| (V) SceneNet [204] | 2017 | 158 | 1×RGB-D | N/A | N/A | Simulated | Simulated | ✓ | ✓ | ✓✓ | ✓ | [205] |
| (V) SUNCG [206] | 2017 | 648 | 1×RGB-D | N/A | N/A | Simulated | Simulated | ✓ | ✓ | ✓✓ | ✓ | [49], [207] |
| ScanNet [208] | 2017 | 823 | 1×RGB-D (custom) | Y | N/A | SLAM | SLAM | ✓ | ✓ | ✓✓ | ✓ | [192], [209] |
| (V) InteriorNet [81] | 2018 | 82 | 1×RGB-D | Y | N/A | Simulated | Simulated | ✓ | ✓ | ✓✓ | ✓ | [49], [56] |
| Bonn [82] | 2019 | 18 | 1×RGB-D | N/A | N/A | MoCap | 3D Scanner | ✓ | ✓ | ✓✓ | ✓ | [82], [210], [211] |
| ETH3D SLAM [212] | 2019 | 39 | 1×color+depth | Y | N/A | MoCap/SfM | N/A | ✓ | ✓ | ✓✓ | | [49], [80], [190], [212], [240] |
| (V) Replica [213] | 2019 | 69 | 1×RGB-D | N/A | N/A | Simulated | Simulated | ✓ | ✓ | ✓✓ | ✓ | [49], [210] |
| VINS-RGBD [215] | 2019 | 14 | 1×RGB-D | Y | N/A | MoCap | N/A | ✓ | ✓ | ✓✓ | | [102], [215] |
| VCU-RVI [67] | 2020 | 0 | 1×RGB-D | Y | N/A | MoCap (P[b]) | N/A | ✓ | ✓ | ✓✓ | | [102], [215], [216] |

[a]MoCap means Motion Capture System, [b]P means partially coverage. Within the tested methods, [48] denotes LSD-SLAM, [49] ORB-SLAM2 (visual and RGB-D), [55] KinectFusion, [56] ElasticFusion, [80] Kintinous, [102] VINS-mono, [146] ORB-SLAM, [190] BundleFusion, [212] BAD-SLAM, [215] VINS-RGBD.

TABLE XX
OVERVIEW AND COMPARISON TABLE-II OF RGB-D-BASED MOBILE LOCALIZATION AND MAPPING AND 3D RECONSTRUCTION DATASETS

| Name | Mobile Platform | Collection Scene | | | Data Sequence | | Challenging | Dataset Link |
|---|---|---|---|---|---|---|---|---|
| | | Indoor | Outdoor | I-O | Number | Volume | | |
| KinectFusion | Handheld | Various | | | 57 | 500-900 frames per seq | | https://hci.iwr.uni-heidelberg.de/benchmarks/KinectFusion_Capture_Tool_and_Example_Datasets |
| TUM RGB-D | Handheld<br>Wheeled Robot | Office, workshop | | | 39 | 1.5m-40m per seq | M | https://vision.in.tum.de/data/datasets/rgbd-dataset |
| 7-Scenes | Handheld | Multiple | | | 14 | 1k-7k frames per seq | | https://www.microsoft.com/en-us/research/project/rgb-d-dataset-7-scenes/ |
| SUN 3D | Handheld<br>Wheeled Robot | Indoors | | | 415 | Tens of thousands of frames in total | | http://sun3d.cs.princeton.edu/ |
| Stanford 3D Scene | Handheld | Multiple | Multiple | | 8 | 72s-432s per seq | | http://qianyi.info/scenedata.html |
| (V) ICL-NUIM | Handheld | Room, office | | | 8 | 2m-11m per seq | | https://www.doc.ic.ac.uk/~ahanda/VaFRIC/iclnuim.html |
| RGB-D Scenes | Handheld | Objects, furniture | | | 23 | 85-1128 frames per seq | | http://rgbd-dataset.cs.washington.edu/dataset.html |
| CoRBS | Handheld | Objects | | | 20 | 5m-59m per seq | M | http://corbs.dfki.uni-kl.de/ |
| SceneNN | Handheld | Rooms | | | 100 | 2k-10k frames per seq | I | http://www.scenenn.net/ |
| ETH RGB-D | MAV | Room | | | 1 | 142s | | https://projects.asl.ethz.ch/datasets/doku.php?id=iros2017 |
| Robot@Home | Wheeled Robot | Rooms | | | 81 | 75min in total | I, F | http://mapir.isa.uma.es/mapirwebsite/index.php/mapir-downloads/203-robot-at-home-dataset |
| (V) SceneNet | Random | Rooms | | | 16k | 5M frames in total | M, I | https://robotvault.bitbucket.io/scenenet-rgbd.html |
| (V) SUNCG | Custom | Rooms | | | 46k | 400k frames in total | | https://sscnet.cs.princeton.edu/ |
| ScanNet | Handheld | Rooms | | | 1513 | 2.5M frames in total | | http://www.scan-net.org/ |
| (V) InteriorNet | Custom | Rooms | | | 15k | 1k frames per seq | M, I, D | https://interiornet.org/ |
| Bonn | Handheld | Office | | | 26 | 332-11k frames per seq | D | http://www.ipb.uni-bonn.de/data/rgbd-dynamic-dataset |
| ETH3D SLAM | Handheld | Various | Multiple | | 96 | 70-4601 frames per seq | M, I | https://www.eth3d.net/slam_datasets |
| (V) Replica | Custom | Rooms | | | Custom | Custom rendered | | https://github.com/facebookresearch/Replica-Dataset |
| VINS-RGBD | Handheld<br>Wheeled Robot<br>Tracked Robot | Lab, Workspace | | | 9 | 32-271s per seq | M, F | https://robotics.shanghaitech.edu.cn/datasets/VINS-RGBD |
| VCU-RVI | Handheld<br>Wheeled Robot | Lab, Hall, Corridor | | | 39 | 3.7km in total | M, I, D | https://github.com/rising-turtle/VCU_RVI_Benchmark |

## D. *Image-based 3D Reconstruction Category*

In this part, we collect 18 datasets that mainly focus on image-based 3D reconstruction usages. The overview and comparison are given in Table XXI and Table XXII.

TABLE XXI
OVERVIEW AND COMPARISON TABLE-I OF IMAGE-BASED 3D RECONSTRUCTION DATASETS

| Name | Year | Cited | Perception Sensor Setup | | | Ground Truth | | Applicable Tasks | | Tested Methods |
|---|---|---|---|---|---|---|---|---|---|---|
| | | | Camera | IMU | LiDAR | Pose | 3D | SfM | MVS | |
| Middlebury [13] | 2006 | 2656 | Professional | N/A | N/A | Robot Arm | 3D Scanner | ✓✓ | ✓✓ | [217], [218], [219], [221] |
| (W) Notre Dame [86] | 2006 | 3899 | Random | N/A | N/A | N/A | N/A | ✓ | ✓ | [86], [229], [232] |
| EPFL [220] | 2008 | 865 | Professional | N/A | N/A | 2D-3D Alignment | 3D Scanner | ✓✓ | ✓✓ | [218], [219], [226] |
| (W) Rome [83] | 2009 | 2150 | Random | N/A | N/A | N/A | N/A | ✓ | ✓ | [83], [86], [90], [219] |
| SAMANTHA [89] | 2009 | 156 | Professional | N/A | N/A | Control Points (P[a]) | 3D Scanner (P) | ✓✓ | ✓✓ | [86], [89] |
| TUM multiview [221] | 2009 | 174 | Professional | N/A | N/A | N/A | 3D Scanner | ✓✓ | ✓✓ | [221], [222] |
| (W) Dubrovnik [85] | 2010 | 457 | Random | N/A | N/A | GPS | N/A | ✓✓ | ✓ | [83], [86], [90], [219] |
| Quad [223] | 2011 | 349 | Professional/ consumer | N/A | N/A | GPS+DGPS | N/A | ✓✓ | ✓ | [86], [87], [90], [225] |
| San Francesco [224] | 2011 | 444 | 6×color (panor[b]) | N/A | 1×Velodyne-64 | GPS+INS | N/A | ✓✓ | ✓ | [87], [225] |
| Tsinghua University [226] | 2011 | 153 | Professional | N/A | N/A | N/A | 3D Scanner | ✓✓ | ✓✓ | [226] |
| (W) Landmarks [87] | 2012 | 383 | Random | N/A | N/A | GPS+INS | N/A | ✓✓ | ✓ | [87] |
| (W) Large-Scale Collections [227] | 2012 | 22 | Random | N/A | N/A | SIFT matching | N/A | ✓✓ | ✓ | [228] |
| (W) SfM-Disambig [229] | 2013 | 58 | Random | N/A | N/A | N/A | N/A | ✓ | ✓ | [229] |
| DTU Robot [230] | 2014 | 146 | Professional | N/A | N/A | Robot Arm | 3D Scanner | ✓✓ | ✓✓ | [219], [231] |
| (W) 1DSfM [232] | 2014 | 259 | Random | N/A | N/A | N/A | N/A | ✓ | ✓ | [86], [90] |
| COLMAP [90] | 2016 | 1306 | Professional | N/A | N/A | N/A | N/A | ✓ | ✓ | [90], [219] |
| ETH3D MVS [233] | 2017 | 245 | 1×Digital<br>2×NA-gray (stereo)<br>2×WA-gray (stereo) | Y | N/A | 2D-3D Alignment | 3D Scanner | ✓✓ | ✓✓ | [90], [219] |
| Tanks and Temples [88] | 2017 | 219 | 1×color (per scene) | N/A | N/A | 2D-3D Registration | 3D Scanner | ✓✓ | ✓✓ | [86], [90], [219] |

[a]P means partially coverage, [b]panor means panoramic. Within the tested methods, [86] denotes Bundler, [90] COLMAP, [219] PMVS.

TABLE XXII
OVERVIEW AND COMPARISON TABLE-II OF IMAGE-BASED 3D RECONSTRUCTION DATASETS

| Name | Mobile Platform | Collection Scene | | | Data Sequence | | Challenging | Dataset Link |
|---|---|---|---|---|---|---|---|---|
| | | Indoor | Outdoor | I-O | Number | Volume | | |
| Middlebury | Robot Arm | Objects | | | 6 | 802 views in total | F, C | https://vision.middlebury.edu/mview/data/ |
| (W) Notre Dame | N/A | | Landmark | | 1 | 715 images | | http://phototour.cs.washington.edu/datasets/ |
| EPFL | N/A | | Buildings | | 12 | 203 images in total | F | https://documents.epfl.ch/groups/c/cv/cvlab-unit/www/data/multiview/index.html |
| (W) Rome | N/A | | Landmarks | | 1 | 16k images | L | https://www.cs.cornell.edu/projects/p2f/ |
| SAMANTHA | N/A | | Landmarks | | 8 | 39-380 images per seq | | http://www.diegm.uniud.it/~fusiello/demo/samantha/ |
| TUM Multiview | N/A | Objects | | | 5 | 21-33 images per seq | F | https://vision.in.tum.de/data/datasets/3dreconstruction |
| (W) Dubrovnik | N/A | | Landmarks | | 1 | 6k images | L | https://www.cs.cornell.edu/projects/p2f/ |
| Quad | N/A | | Landmarks | | 1 | 6k images | L | http://vision.soic.indiana.edu/projects/disco/ |
| San Francesco | Vehicle | | Landmarks | | 1 | 1.7M images | L | https://purl.stanford.edu/vn158kj2087 |
| Tsinghua University | N/A | | Buildings, temples | | 9 | 68-636 images per seq | T | http://vision.ia.ac.cn/data/ |
| (W) Landmarks | N/A | | Landmarks | | 10k | massive | L | http://landmark.cs.cornell.edu/Landmarks3/ |
| (W) Large-Scale Collections | N/A | | Landmarks | | 5 | 4k-7k images per seq | L | https://research.cs.cornell.edu/matchlearn/ |
| (W) SfM-Disambig | N/A | | Landmarks | | 4 | 839-8032 images per seq | L | https://www.cs.cornell.edu/projects/disambig/ |
| DTU Robot | Robot Arm | Objects | | | 80 | 49/64 views per seq | F, C | http://roboimagedata.compute.dtu.dk/?page_id=36 |
| (W) 1DSfM | N/A | | Landmarks | | 13 | 227-2152 images per seq | L | https://research.cs.cornell.edu/1dsfm/ |
| COLMAP | N/A | Landmarks | Landmarks | ✓ | 4 | 100-1273 images per seq | L | http://colmap.github.io/datasets.html |
| ETH3D MVS | Tripod/Handheld | Various | Multiple | | 82 | 11k images in total | F | https://www.eth3d.net/datasets |
| Tanks and Temples | Stabilizer | Various | Various | | 14 | 4k-22k images per seq | F, C, L | https://www.tanksandtemples.org/ |

## VI. Evaluation Criteria

Although reliable ground truth are the decisive premise, evaluation criteria are also indispensable for benchmarks. They play important roles in correctly quantifying and fairly comparing the algorithm performances, identifying potential failure modes, and thus looking for breakthroughs. In this section, we will focus on 2 indicators: positioning and mapping, which are the main functionalities of SLAM related problems. Some other non-functional concerns like time consuming, CPU&RAM usage, and GPU requirement, although have been investigated by some efforts [234], [235], [236], will not be covered due to the immaturity and poor utilization. In part *C*, we compare top results on 2 different benchmarks to illustrate the necessity of complementary usage.

### *A. Positioning Evaluation*

Due to the limitation of large-scale surveying techniques in earlier years, it was rather difficult to obtain ground truth 3D maps. Therefore, it has been a long while to evaluate SLAM related algorithms by examining the positioning performance. The most widely used evaluating indicators could be the two proposed within the TUM RGB-D benchmark – relative pose error (*RPE*), and absolute trajectory error (*ATE*) [12].

The relative pose error (RPE) investigates the local accuracy of the motion estimation systems across a fixed time interval Δ. If the estimated trajectory poses are defined as $\mathbf{P}_1,\ldots,\mathbf{P}_n \in \mathrm{SE}(3)$ and the ground truth poses are defined as $\mathbf{Q}_1,\ldots,\mathbf{Q}_n \in \mathrm{SE}(3)$, then the RPE at time step $i$ could be defined as:

$$\mathbf{E}_i := (\mathbf{Q}_i^{-1}\mathbf{Q}_{i+\Delta})^{-1}(\mathbf{P}_i^{-1}\mathbf{P}_{i+\Delta}) .$$

Considering a sequence of $n$ camera poses, there could be $(m = n - \Delta)$ individual RPEs. Then how to measure the performance on the whole trajectory? TUM RGB-D benchmark proposed to compute the root mean square error (RMSE) over all time intervals of the translational component as:

$$\mathrm{RMSE}(\mathbf{E}_{1:n},\Delta) := \left(\frac{1}{m}\sum_{i=1}^{m}\left\|trans(\mathbf{E}_i)\right\|^2\right)^{1/2} ,$$

where the $trans(\mathbf{E}_i)$ refers to the translational component of the RPE. Note that due to the amplification effect of the Euclidean norm, the RMSE could be sensitive to the outliers. So if desired, it is also reasonable to evaluate the mean or median errors as an alternative. Although only the translational errors are measured here, the rotational errors can also be reflected together inside. Additionally, the selection of time interval Δ could vary with different systems and conditions. For instance, the Δ can be set to 1 to examine real-time performance, resulting as per frame drift; as for systems inferring via several previous frames, performing local optimization, and focusing on large-scale navigation, it is not necessary nor reasonable to count on the estimation of each individual frame. However, it is not appropriate to set Δ to $n$ directly, because it penalizes early rotational errors much more than the occurrences near to the end. Instead, it is recommended to average the RMSEs over a wide range of time intervals, but to avoid excessive complexity, an approximation that counts on a fixed number of intervals could also reach a compromise.

Absolute trajectory error (ATE) measures the global accuracy by comparing the estimated trajectory against the ground truth to get absolute distances. As the 2 trajectories could lie in different coordinates, an alignment via a rigid-body transformation **S** that maps the estimated poses $\mathbf{P}_{1:n}$ onto the ground truth poses $\mathbf{Q}_{1:n}$ is prerequisite. Then the ATE at time step $i$ could be computed as:

$$\mathbf{F}_i := \mathbf{Q}_i^{-1}\mathbf{S}\mathbf{P}_i .$$

Similar with RPE, the root mean square error (RMSE) over all time indices of the translational component is also proposed as one possible metric for ATE:

$$\mathrm{RMSE}(\mathbf{F}_{1:n}) := \left(\frac{1}{n}\sum_{i=1}^{n}\left\|trans(\mathbf{F}_i)\right\|^2\right)^{1/2} .$$

It should be noted that ATE considers only the translational errors, but actually the rotational errors will also result in wrong translations. Compared with RPE, ATE has an intuitive visualization for inspecting the actual accidental area of the algorithm. An example visualization of ATE and a comparison between ATE and RPE are shown in Fig 21. As shown in the comparison, RPE could be slightly larger than ATE due to the more impact from rotational errors.

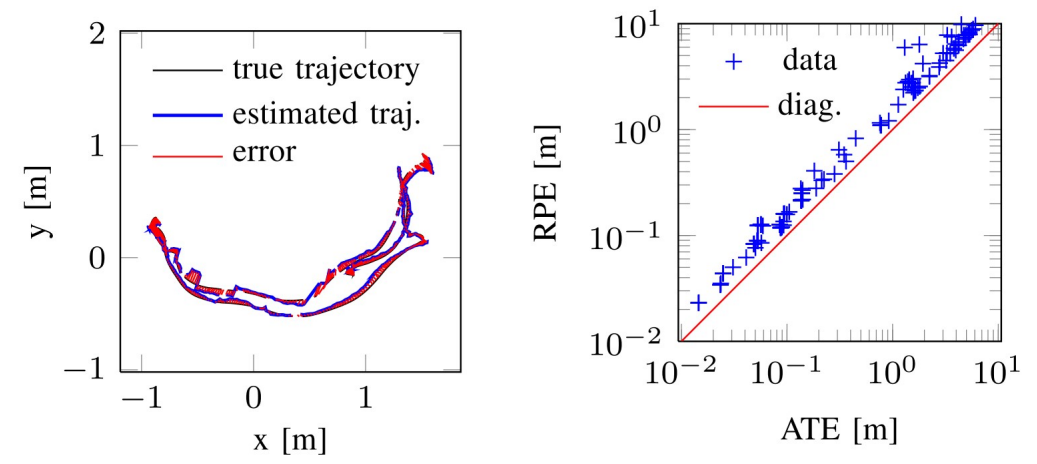

Fig. 21. An example visualization of ATE on the "fr1/desk2" sequence of the TUM RGB-D Dataset (left) and a comparison between RPE and ATE (right). Both plots were generated by RGB-D SLAM system [12].

There are many mainstream datasets [126], [41], [134] sharing the same evaluation criteria with the above, or developing some extended versions based on the same mechanism. For instance, KITTI Odometry benchmark[21] extends RPE to study rotational errors also to enable deeper insights; TartanAir defines success rate (SR) as the ratio of non-lost frames against total sequence length, as the lose-trackings cannot be calculated RPE and ATE.

[21] http://www.cvlibs.net/datasets/kitti/eval_odometry.php

However, as we have mentioned in Section II, not all datasets are able to provide complete ground truth trajectory [126], [38]. In this case, normally the start and end segments are available, then after alignment, still RPE and ATE can be evaluated.

### B. *Mapping Evaluation*

In the existing SLAM algorithm publications, it is quite unusual to come across evaluations on mapping performance. But indeed, the evaluation criteria on 3D reconstruction have existed for a long time, which mainly originated from the Middlebury dataset [13]. Denoting the ground truth geometry as $G$ and the reconstruction result as $R$, there are generally two indicators to be considered: accuracy, and completeness.

The accuracy measures how close $R$ is to $G$ by computing the distances between the corresponding points, which are usually determined by finding the nearest match. If the reconstructions or ground truth models are in other formats like triangle meshes, then the vertices could be used for comparison. One issue will be encountered where the $G$ is incomplete, resulting the nearest reference points to fall on the boundary or a distant part. In this case, a hole-filled version $G'$ is proposed for computing the distance, and the points matched to the repaired region will be discounted. Fig. 22 well illustrates the principle.

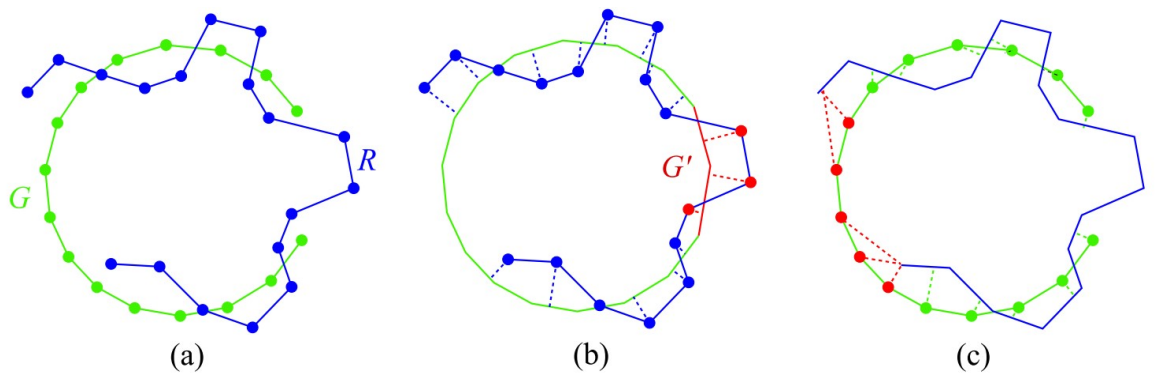

Fig. 22. Evaluation of the reconstruction result against the ground truth model. (a) the representation of the two models, both with incomplete surface. (b) to measure the distance, the nearest matches are found, and the matches related to the hole-filled area will not be included. (c) to measure the completeness, the distances between point matches are computed, those run beyond the threshold will be regarded as “not covered” [13].

The completeness investigates how much of $G$ is modeled by $R$. Opposite from the accuracy indicator’s comparing $R$ against $G$, the completeness measures the distances from $G$ to $R$. Intuitively, if the matching distance runs beyond a threshold, we can deem there is no corresponding point on $R$, hence the point on $G$ can be logically regarded as “not covered”, as illustrated in Fig. 22.

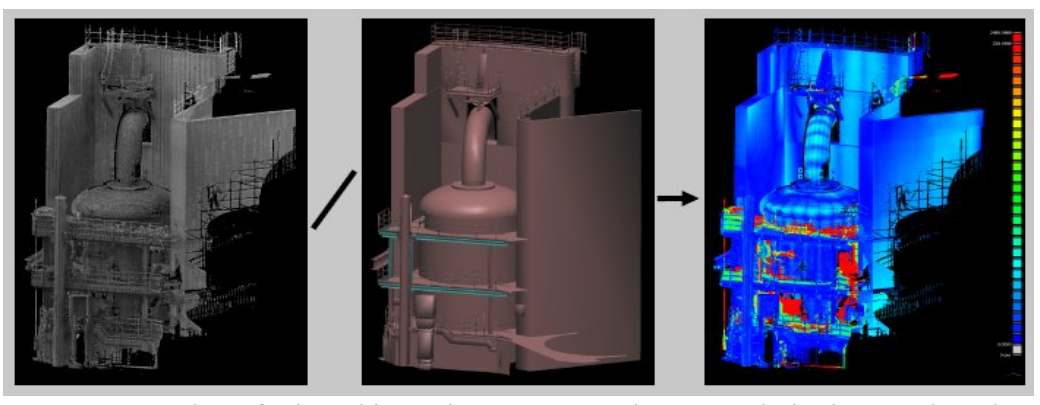

Fig. 23. An example of cloud/mesh computation module investigating the reconstruction accuracy in CloudCompare software [237], the different colors represent different distance values.

For the convenience of the evaluation, ICL-NUIM proposed to exploit the open-source software CloudCompare[22] to compute the distances between the reconstruction and the ground truth models. Five standard statistics can be obtained: Mean, Median, Std., Min., and Max. An evaluation demo is shown in Fig. 23.

[22] https://www.danielgm.net/cc/

### C. *Complementary usage of datasets*

As shown in Section V, SLAM related datasets can vary in many dimensions, which cause different impacts (motion pattern, scene appearance and structure, etc.) on algorithm performance and lead to many challenging effects. However, as revealed by the citation statistics, the usages of the datasets are quite biased, which will potentially lead to over-fitting on certain benchmark and hidden defects. To illustrate such risk effectively, here we choose the most widely used KITTI odometry benchmark (only vision-based methods shown) and UZH-FPV benchmark (results of IROS 2020 competition) to give a comparison on the TOP 7 results using the RPE metric proposed by KITTI benchmark (as shown in Table XXIII and Table XXIV). The 2 datasets are with totally different platforms, scenes, sequences, and with a huge contrast on citations (KITTI 6341 *vs*. UZH-FPV 47).

TABLE XXIII
THE BENCHMARK RESULTS ON KITTI DATASET (VISION ONLY) [73]

| Ranking | Name | Translation (%) | Rotation (deg/m) |
|---|---|---|---|
| 1 | SOFT2 | 0.57 | 0.0010 |
| 2 | SOFT-SLAM | 0.65 | 0.0014 |
| 3 | RADVO | 0.82 | 0.0018 |
| 4 | LG-SLAM | 0.82 | 0.0020 |
| 5 | Rotrocc++ | 0.83 | 0.0026 |
| 6 | GDVO | 0.86 | 0.0031 |
| 7 | SOFT | 0.88 | 0.0022 |

TABLE XXIV
THE BENCHMARK RESULTS OF THE UZH-FPV DATASET [71]

| Ranking | Name | Translation (%) | Rotation (deg/m) |
|---|---|---|---|
| 1 | MEGVII-3D | 6.819 | 0.263 |
| 2 | VCU Robotics | 6.891 | 0.265 |
| 3 | LARVIO | 6.919 | 0.266 |
| 4 | Lenovo_LR_ShangHai | 7.005 | 0.282 |
| 5 | Basalt | 7.494 | 0.268 |
| 6 | Xin Zhang | 9.140 | 0.219 |
| 7 | QuetzalC++ | 34.273 | 1.635 |

As can be seen, the TOP results on the two benchmarks have a huge contrast: compared with KITTI, the translational error on UZH-FPV benchmark is one order of magnitude larger, and the rotational error is two orders of magnitude larger. This happens even in the context that the methods from UZH-FPV all fuse the IMU data, whereas the listed methods from KITTI only use pure vision data. Of course, not all the datasets are going to form this big of a contrast on algorithm performance, but it is a truth that the results will vary. Thus, from this point of view, one should not be overconfident on a certain benchmark, since many defects could be hidden. But this never means the datasets produce good scores should be given up: on the one hand, the mechanisms of algorithms indeed vary, and one cannot ensure the algorithm will not encounter certain challenges in “easier” datasets (especially for learning-based methods); on the other hand, from the application point of view, the results in different scenarios all make sense. To this end, we do not encourage to draw biases on dataset selection based on benchmark results (most datasets even have not been tested and evaluated comprehensively after all), instead, we should use the datasets in a complementary manner and take advantage of the unbalanced results to look for breakthroughs.

## VII. Analyses And Discussions

As mentioned repeatedly in the paper, SLAM related datasets complement each other and together form the dataset community. Thus except for understanding the datasets and their difference, to study the whole makes another big sense. Therefore, based on the overview and comparison in Section V, this section studies the whole and give the current limitations and future directions of datasets and evaluation criteria. In the end, we discuss about how to build comprehensive datasets for SLAM community.

### *A. Datasets limitations and directions*

For the datasets limitations and directions, we mainly choose 7 dimensions to analyze and discuss, respectively on the 5 fundamental dimensions including mobile platform, sensor setup, target scene, data sequence, and ground truth, and 2 hot topics recently namely challenging elements, and virtual and synthetic datasets.

#### *1) Mobile platform*

For current limitations, the first is the deficiency in the variety of mobile platforms. Fig. 24 shows the distribution of different mobile platforms in each dataset category (we have not shown image-based category because normally good image quality are ensured while collection, e.g., using a stabilizer, thus the motion pattern will not affect too much). As can be seen, vision-based category has a relatively balanced distribution, whereas LiDAR-related category mainly draws on car and wheeled robot (both with 3-DoF motion) and RGB-D-based category mainly draws on handheld and wheeled robot. Although this is partly resulted by the application scenario, from the algorithm developing point of view, the deficiency in the variety of motion patterns will potentially hide the defects. Another limitation is that, most of the datasets only employed single mobile platform for collection. Although they can be used in a complementation way, it is really unfair to make horizontal comparisons if one want to investigate the impacts of different motion patterns.

For future directions, firstly, more diverse platforms are encouraged, e.g., boat [119], motorcycle [123], and backpack [177]. Secondly, we expect more 6-DOF platforms in LiDAR-related datasets, e.g., UAV and backpack [9] are increasingly used in mobile surveying and badly in need of validation. Thirdly, it is highly recommended to deploy various platforms in each single scenario and route to enable fair horizontal comparisons.

#### *2) Sensor setup*

Considering that the mainstream sensors can be effectively calibrated temporally and spatially, the limitation of sensor setup mainly draws on the deficiency of number and types. Leveraging multiple sensors and modalities fusion is an inevitable trend to improve system accuracy and reliability, especially stereo-vision and IMU fusion methods, which have been widely employed and proved to be effective. However, as shown in Section V, many datasets only use basically required sensors: TUM MonoVO [38] only equips monocular cameras, and most of the RGB-D datasets do not have IMU data. Moreover, at present, event-based vision is earning rapid developments and playing a crucial role in robust and high-speed robotics, while are suffering from acute shortage in related datasets, in need of more dedication.

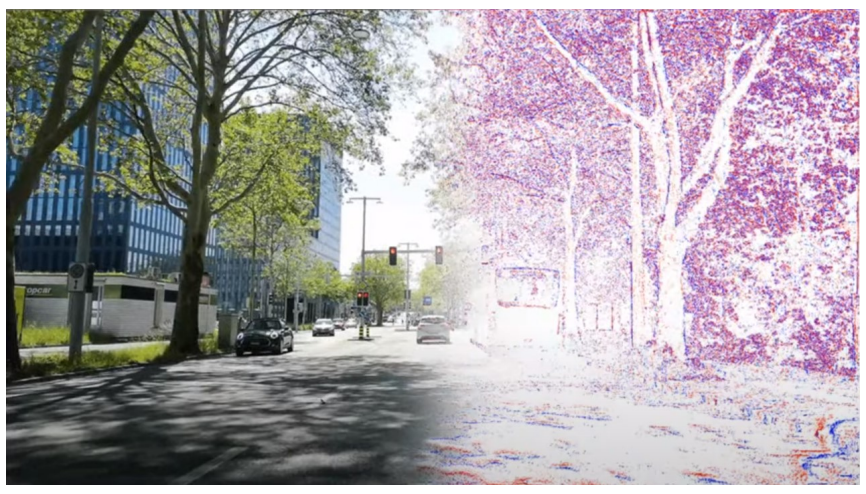

Fig. 25. An appoach to convert existing vision datasets to events [238].

In the future, we encourage every dataset to equip redundant and diverse sensors (e.g., event and thermal), to enable possible development and comparison. Then specifically, we hope every future dataset to setup at least stereo cameras and an IMU, which are ubiquitous and affordable. In addition, more dedicated works are needed for event datasets, not only real-world collections, but also convert normal vision data to events (see Fig. 25).

#### *3) Collection scene*

For current limitations, firstly, there is still a deficiency in the variety of scene type. For example, only 3 datasets [41], [112], [187] have covered shopping mall, which is highly dynamic and quite common in real life. Secondly, there is a serious lack of indoor-outdoor integrated scenes. Whereas in daily life, it is quite usual for a robot or a human to traverse back and forth. Thirdly, datasets with large-scale indoor scenes are still not enough. Most of the interior scenes are office and room, which cannot meet the demands of applications like factory robots and BIM.

Accordingly, for future directions, firstly, more diverse scenes are expected to complement current datasets, especially typical scenarios like shopping mall, ruins, and industrial park. Secondly, more indoor-outdoor integrated scenes are urgently demanded to research seamless navigation. Thirdly, more large-scale indoor scenes are expected, such as big buildings and large factory hall.

#### *4) Data sequence*

The limitation of data sequence mainly draws on the shortage of sequence number and variety and length of single session. For example, Zurich Urban [111] only collected single sequence in large-scale urban environment; ICL-NUIM [79] only traversed 2m-11m's length per session, which is even too short compared with other indoor datasets. Such datasets have not made full use of the scene and wasted the chance of more thorough testing.

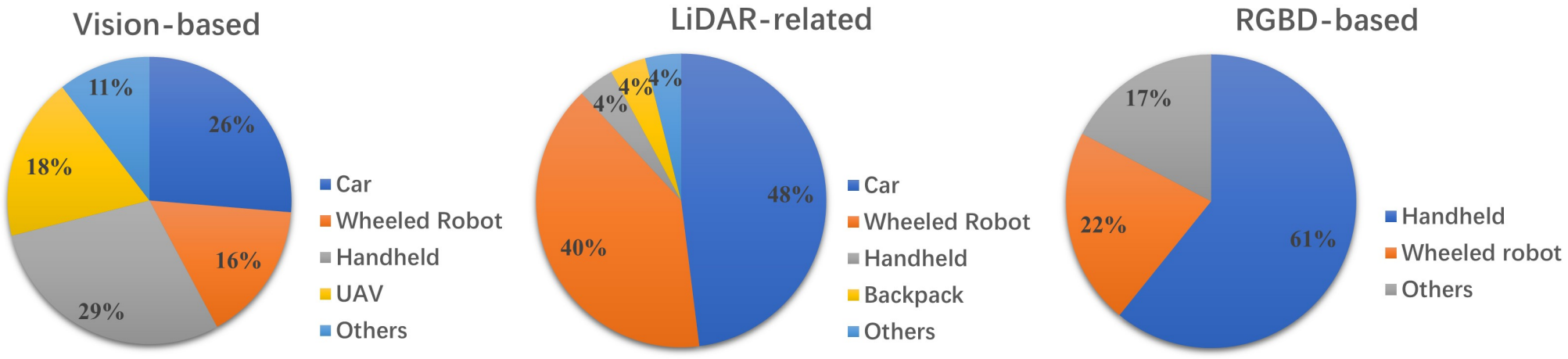

Fig. 24. The distribution of different mobile platforms in each dataset category (except for image-based).

To this end, making full use of the scenes is expected in future datasets, namely collecting as many as possible data sequences and traverse various and longer routes. This is especially doable in large-scale environments. Besides, we expect every dataset to post the session length statistic.

*5) Ground truth*

For pose ground truth, the limitation mainly lies on the low accuracy and incomplete coverage in some datasets. For example, Complex Urban [37] (in outdoor GNSS-denied areas) has to use SLAM to generate a baseline; TUM VI [126] (long-range indoor and outdoor integrated) only has ground truth data in the start and end segments. For 3D ground truth, the primary limitation is that most datasets (62/97) have not provided such data, and there are also some datasets using SLAM-based algorithm to generate 3D baselines, which could be less accurate.

In the future, we expect there could emerge novel techniques and approaches for generating pose ground truth, especially those with versatility both indoors and outdoors. For 3D structure, we expect future datasets to provide ground truth as far as possible, to enable mapping evaluation. Besides, more types of ground truth are expected, such as depth image and semantic labels.

*6) Challenging elements*

To date, there has already been many challenging elements identified and embedded in the datasets, proven to shake the performance of algorithms (Fig. 26 shows some failure cases on TartanAir [72]). But there is still a deficiency in some expressly designed elements, such as dense dynamics, multiple time slots, extreme weather, and different seasons. For example, in RGB-D category, only InteriorNet [81] and Bonn [82] have designed sequences with moving objects, whereas such situation are quite commonly encountered in real-world deployments.

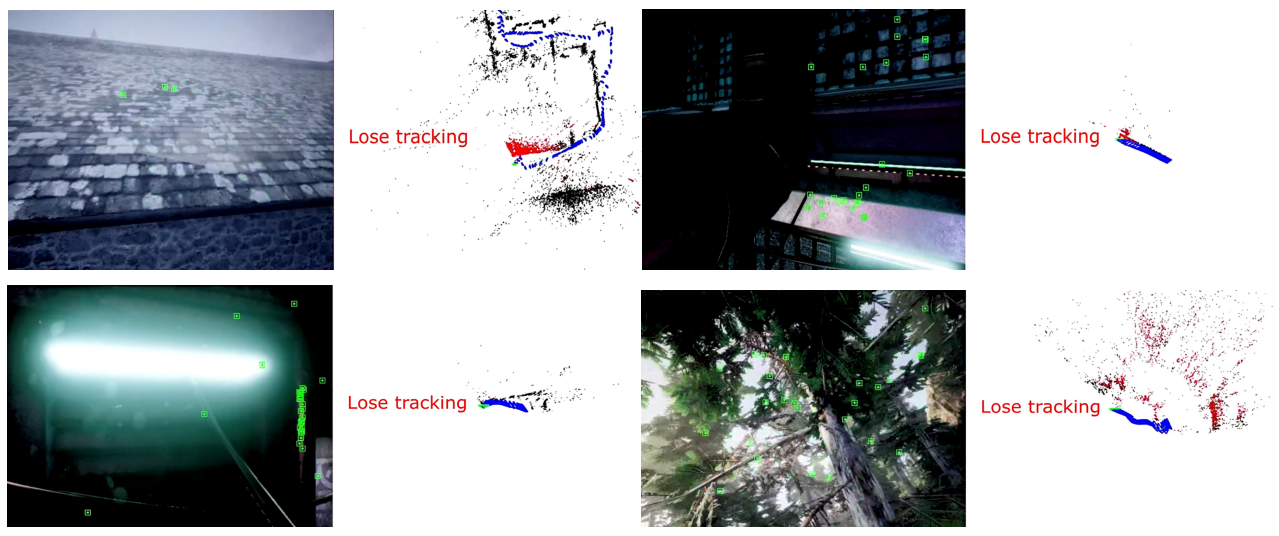

Fig. 26. Some failure cases of ORB-SLAM-Stereo under challenging conditions such as weak and repetitive texture, and bad illumination [239].

In the future, firstly, more and more datasets with dense and various challenging elements are expected, especially those as mentioned above. Then, rigorous studies on the failure modes of algorithms are highly demanded, which will help to quantify the impacts of challenges and identify more such elements.

*7) Virtual and synthetic datasets*

Synthetic datasets open a new route to handle the drawbacks of real-world datasets, such as complicated collection procedure, limited scenes and sequences, and inaccurate ground truth. For such datasets, the limitation mainly draws on the deficiency of data reality. Firstly, the rendered images are still far from realistic sometimes [72], [92]. Secondly, it is still quite rare to mimic real-world sensor latency and noises, e.g., camera motion blur and LiDAR motion distortion, which will make the data too ideal.

Therefore, in the future, considering that synthetic datasets can generate boundless sequences under various scenes, especially with perfect ground truth, more such datasets are encouraged to bridge certain gap, but with higher data reality.

## *B. Evaluation limitations and directions*

For limitations and directions of evaluation criteria, we mainly choose 3 dimensions to analyze and discuss, respectively on the 2 functional dimensions namely localization and mapping, and 1 non-functional dimension namely hardware consumption.

*1) Localization evaluation*

With the wide implementations and test of time in past years [240], [241], [242], the evaluation criteria on localization are relatively mature, which mainly consists of ATE, RPE, and the related variations. However, such dimensions are still quite limited, because SLAM tasks contain many processes, such as initialization, loop-closure, and relocalization. To this end, in the future, more diversified and detailed indicators are expected, such as the accuracy and speed of initialization, the accuracy and success rate of relocalization, the detection rate of loop-closure [41], and the localization accuracy under challenging conditions [243]. These dimensions will definitely be of great significance for inspecting an algorithm deeply and all-roundly.

*2) Mapping evaluation*

As reflected in the literature, mapping performance is rarely measured in SLAM, whereas sometimes, building an accurate map can even be more important than localization, e.g., path planning, barriers avoidance, and making high-definition (HD) map and light-weight semantic map [144], [244] as virtual sensor for localization. Besides, the existing criteria only investigate the 3D geometry [79], [13], overlooking other dimensions. Hence in the future, the algorithms should be widely evaluated in mapping, and novel criteria that combine geometry, texture, and semantic labeling are desiderated by the community.

*3) Hardware consumption evaluation*

The current evaluation criteria and algorithms mainly focus on the functional level – localization and mapping, overlooking the hardware consumption, which is a sharply decisive part for the transition from laboratory to application. Thus in the future, more such criteria are expected to be proposed and especially widely employed, such as CPU and RAM usage, GPU requirement, and time consumption [234]. These indicators determine whether an algorithm can be widely deployed in real world, namely, whether can be embedded lightweightly, whether can work in real time, and whether the total cost is reasonable.

## *C. How to build comprehensive datasets?*

Comprehensive datasets are the powerhouse of improvements and breakthrough of algorithms. Before the release of ImageNet in 2009 [20], no one had imaged a vision Artificial Intelligence can be like today within 10 years' trip: The accuracy of objects classifying rose from 71.8% to 97.3%, surpassing human ability and proving that datasets can push the limits of algorithms to another level [245]. However, compared with machine learning, to collect SLAM datasets is more complicated and time consuming, thus it is almost impossible to build an ImageNet in SLAM field.

In such condition, we mainly advocate 2 approaches to bridge the gap. One is to develop mature techniques and pipeline to generate high-reality synthetic datasets. This is exactly what many datasets are trying hard to do, such as Replica (using Habitat-Sim) [213], [246] and InteriorNet [81]. Moreover, surprisingly, from RTX20 series, NVIDIA has released their ray-tracing GPUs with RT Cores [247]. Such hardware-level ray-tracing technology enables users to get high-quality photo renderings in real-time (see Fig. 27, a screenshot of Red Dead Redemption 2 game rendered by RTX3090), leading to a more convenient and boundless way to generate realistic SLAM datasets.

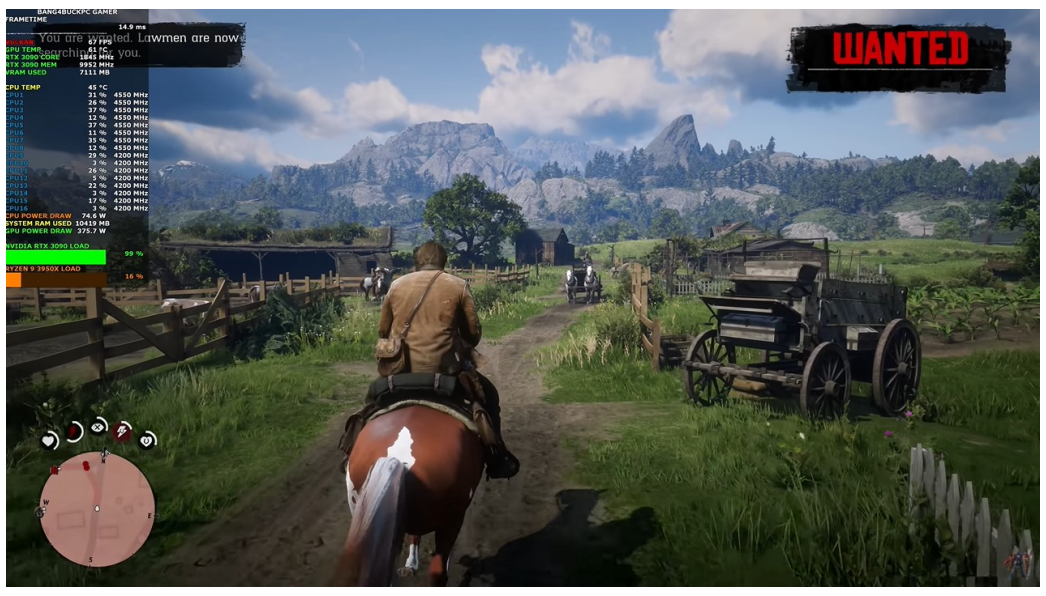

Fig. 27. A screenshot of Red Dead Redemption2 rendered by RTX3090 [248].

The other is to build a set of standards of datasets and call for distributing the data collections (e.g., different platforms, scenes, and weathers, etc.) to the community, and finally the standard-compliant datasets are released to public. In this way, both individuals and institutes can join collections and enjoy the feedbacks.

It is worth noting that, it is a matter of principle that simulation cannot replace experiment of real-world data, especially for tasks like SLAM, which are close related to life safety issues. Thus it is recommended to use them jointly: Use the synthetic datasets as preliminary testing purposes to validate a certain assumption, then after getting supportive results, the corresponding real-world datasets can be proposed. We believe in the next few years, the datasets community can witness a prosperity and drive SLAM technology significantly.

## VIII. Conclusion And Future Work

This paper is the first comprehensive survey specialized on SLAM related datasets. We treat SLAM related dataset as a research branch rather than merely a tool, thus except for the normal review dimensions of datasets, we also introduce the collection methodology, which distinguishes our survey from many others. By presenting a range of cohesive topics, two major concerns have been addressed: one is to remove the biased usage of datasets and ensure a comprehensive, correct, and efficient selection; the other is to lower the threshold and promote future directions of datasets research. We believe by making full use of the existing datasets and exploiting new ones, SLAM related algorithms can attend key breakthroughs soon afterwards. Moreover, to maintain the benefit of the paper, we will keep updating the datasets dictionary and comments periodically on GitHub at: https://github.com/robot-pesg/SLAM_Datasets_Survey_JAS.

The paper itself is already comprehensive and concrete which well accomplishes its missions about datasets, but afterwards, another key topic arises namely the study of "dataset-algorithm-evaluation". Although many qualitative challenging elements are pointed out in each dataset, how to quantify such elements? What are the quantitative impacts of them on algorithms? We do not want to crowd this survey too much which may cause emanative structure, instead we believe such issues deserve a self-contained research paper, which is exactly our main goal in the coming few months, attending to thorough evaluations of diverse algorithms against massive datasets. Besides, we are also building a new vision-LiDAR dataset in typical, large-scale, and indoor-outdoor integration scenes with both accurate localization and mapping ground truth.

## Acknowledgment

To achieve a better illustration, this survey has reused many photos from the internet, dataset holders, and journals. The authors would like to acknowledge them all for the public licenses and copyright permissions. The authors would also like to acknowledge Dr. Yassine Selami for the refinement on English writing. Yuanzhi Liu would like to thank Prof. Bart Goossens and Prof. Wilfried Philips for the guidance and suggestions on the research when he served as a guest researcher in Ghent University, Belgium.